\documentclass{article}
\usepackage[utf8]{inputenc}
\usepackage[T1]{fontenc}
\usepackage{graphicx}
\usepackage{amsmath}
\usepackage{amssymb}
\usepackage{hyperref}
\usepackage{multirow}
\usepackage{makecell}
\usepackage{booktabs}
\usepackage[a4paper, margin=1in]{geometry}
\usepackage{algorithm}
\usepackage{algpseudocode}
\usepackage{tikz}
\usetikzlibrary{arrows.meta}
\usepackage{subcaption}
\usepackage[affil-it]{authblk}
\usepackage[numbers,sort&compress]{natbib}

\usepackage[table]{xcolor}
 \definecolor{hlA}{HTML}{0072B2}
 \definecolor{hlB}{HTML}{D55E00}

\newcommand{\greyl}[1]{\textbf{\textcolor{hlA}{#1}}}
\newcommand{\redl}[1]{\textbf{\textcolor{hlB}{#1}}}

\title{MOTe error: A decomposable framework for evaluating Document Layout Analysis models}

\author[1,2]{Jonathan Bourne\thanks{Corresponding author: \href{mailto:jonathan@the3tc.ai}{jonathan@the3tc.ai}}}
\author[3]{Mwiza Simbeye} 
\author[4]{Ishtar Govia}

\affil[1]{The 3TC AI, London, United Kingdom}
\affil[2]{Centre for Advanced Spatial Analysis, University College London, London, United Kingdom}
\affil[3]{Independent researcher}
\affil[4]{Amagi Brain Health, London, United Kingdom}

\date{}

\begin{document}

\maketitle

\begin{abstract}
Document Layout Analysis (DLA) is a machine vision task that parses a page into meaningful elements, and is a critical stage in Document Understanding (DU). DLA models are usually evaluated with IoU-based metrics such as the F1. However, these metrics are designed for images that are 2D projections of 3D space, not for the natively 2D imagery of printed media. We show both theoretically and empirically that the F1 suffers from two critical failures in the context of DU. These failures are granularity sensitivity, which can lead to F1 collapse even with perfect text extraction; and low semantic competence, which can result in high F1 scores that mask substantial parsing-induced character errors. To remedy these failures we introduce: the Structural Semantic Unit (SSU), a relational labelling approach that shifts the focus from the physical to the semantic structure of the content; and Missing, Overlap, Trespass, and Excess (MOTe) error, a decomposable metric for measuring page parsing error. MOTe correlates more strongly with parsing-induced Character Error Rates (CER) than the F1 ($|\rho|$ = 0.82 vs 0.54).
MOTe also yields idealised upper and lower bounds on CER that hold in 99\% and 82\% of model predictions respectively. The lower bound can act as a direct proxy for CER with an $R^2=0.84$.
We found that given two models MOTe picks the one that would produce the lowest CER 86\% of the time, outperforming pixel-level Dice (77\%) and F1 (68\%). Finally, we release a Python library for applying MOTe in DLA projects.
\end{abstract}

\maketitle
\section{Introduction}
\label{sec:intro}

Document Understanding (DU) is the process by which printed media is converted into computer readable format. A critical stage of Document Understanding is Document Layout Analysis, which focuses on parsing a page into conceptually distinct regions, ready for efficient data extraction and digitisation. A key part of this process is ensuring that parsing-induced character errors are minimised. These errors occur when the parsing model misses, duplicates, or combines unrelated parts of the page, resulting in missing, repeated or mis-ordered text. This can affect subsequent text analysis and can complicate the evaluation of processes such as Optical Character Recognition.

Document Layout Analysis has developed rapidly in the last few years, driven by advances in the transformer architecture and the resulting AI boom. However, the metrics most commonly applied to printed media are the same as those used in general machine vision, which are the F1, mean Average Precision (mAP), and Intersection over Union (IoU).

The lack of comprehensive evaluation metrics specific to Document Layout Analysis is important, as this field is fundamentally different to the scenes commonly captured in photography and the competitions that initially pushed machine vision forward \cite{lin_microsoft_2015, everingham_pascal_2010}. For example, as photographs are 2D projections of 3D space, it is common for them to have objects that overlap (people in a crowd), interact (dog holding frisbee), and images that are sparse (two people waving at each other). In contrast, printed media, which is natively 2D and explicitly designed for human comprehension, is functionally an irregular tessellation which attempts to minimise the empty space without impacting readability. 

When framed in such terms, the metrics that work so well in natural scenes appear lacking in crucial information. Do the predictions overlap? How much of the page are they actually covering? Do the spatial boundaries of the predictions align with the semantics of the text? To resolve this previous authors such as  \cite{shafait_performance_2008} and \cite{clausner_scenario_2011}, introduced more expressive error profiling methods that introduced new and valuable concepts such as splits (ground truth covered by more than one predictions) and merges (predictions cover more than one ground truth), allowing the evaluation of errors that create parsing-induced character errors. However, these systems were complex (\cite{clausner_scenario_2011} had up to 800 weights), did not provide a clear mathematical framework, used thresholds \cite{shafait_performance_2008}, and were explicitly profilers rather than metrics. Building on such earlier work, \cite{heo_led_2026}, provide additional concepts, such as hallucination (predictions where there is no content) and duplicate (highly overlapping predictions), along with mathematical definitions. However, this approach is designed to test the abilities of Large Language Models to diagnose and reason about structural errors as opposed to being a metric for page parsing itself.

What is clear is that the continued dominance of the mAP and F1 based approaches \cite{jaume_funsd_2019, zhong_publaynet_2019, auer_icdar_2023, cheng_m6doc_2023, sun_pp-doclayout_2025,  zottin_icdar_2025} show that no evaluation framework, distinct from the IoU and specifically designed for 2D native imagery, has achieved widespread adoption. As such, Document Layout Analysis practitioners are left with generalist machine vision methods, which are not sufficiently expressive to provide meaningful insight into model performance, leaving them to use qualitative approaches, for debugging and analysis. 

Although evaluation metrics are designed to measure the dyadic relationship between a model and the ground truth, this itself can be a source of problems. A label is only correct relative to the schema that defined it, making classification, including defining objects, highly subjective. This can be thought of in terms of Wittgenstein's concept of "meaning as use", as a dataset is labelled according to the goals and assumptions of the practitioner designing it, not according to some set of universal rules, a view shared by earlier DU researchers \cite{shafait_performance_2008, clausner_scenario_2011}. 

Document labelling schema are an example of this subjectivity, PAGE \cite{pletschacher_page_2010}, META/ALTO \cite{stehno_metaeautomated_2003}, hOCR \cite{breuel_hocr_2007}, TEI \cite{tei_tei_2025}, and ABBYY FineReader all have overlapping, but not entirely interoperable, differences. The situation is exacerbated by popular datasets \cite{zhong_publaynet_2019, cheng_m6doc_2023, pfitzmann_doclaynet_2022}, which often use their own custom labelling schemas that may not be fully described.

As a result the labelling of any one dataset is at best internally coherent \cite{northcutt_pervasive_2021,  beyer_are_2020, whang_data_2023}, but there is no guarantee of it being compatible with another, apparently, similar dataset \cite{paullada_data_2021, bevandic_automatic_2022, meng_detection_2023, jalocha_label_2025, hussain_quality_2026}. 

Beyond the most immediate issue of mismatched classes, this can be a problem for evaluating DLA models as IoU, on which the F1 and mAP are calculated, is sensitive to the arbitrary choice of the granularity at which text is labelled (e.g. line or paragraph). Large differences in prediction and ground truth granularity causes the IoU to collapse giving the appearance of poor performance even under perfect parsing, due to superficial splits and merges \cite{clausner_scenario_2011, heo_led_2026}.

The lack of expressivity and the labelling issues exemplified by granularity mismatch, suggest the relationship between data, model, and metric forms a triad, and is richer than a simple dyad between predictions and ground truth (see Figure \ref{fig:triad_pragmatic}). Labelling differences between train and test data can be seen as degrading this triad, affecting the ability to evaluate without model fine-tuning. We borrow the term `pragmatic failure' \cite{thomas_cross-cultural_1983} from semiotics, to describe a metric that cannot interpret a model within the context of a given test data, because it is like a failure of understanding in a spoken conversation. Similarly we operationalise `pragmatic competence', such that, a more pragmatically competent metric is one whose scores track task-relevant outcomes (e.g. page parsing quality) across a wider range of granularities and schemas, without requiring specific retraining of the model.

\begin{figure}
    \centering
        \resizebox{0.5\linewidth}{!}{
\begin{tikzpicture}[
    box/.style={
        draw=black!80,
        rounded corners=6pt,
        fill=white,
        minimum width=2.4cm,
        minimum height=1.0cm,
        font=\large\bfseries,
        line width=1.2pt
    },
    darrow/.style={
        <->,
        line width=1.0pt,
        >=Stealth,
        shorten >=4pt,
        shorten <=4pt
    }
]

\node[font=\large\bfseries, align=center]
    at (0, 4.5) {The triad of data, model, and metric};

\node[box] (data)   at (0, 3.2)  {Data};
\node[box] (model)  at (-3.0, 0) {Model};
\node[box] (metric) at (3.0, 0)  {Metric};

\draw[darrow, dashed] (model)  -- (metric);
\draw[darrow] (data)   -- (model);
\draw[darrow] (data)   -- (metric);

\end{tikzpicture}
}
\caption{The relationship between the data, model, and metric. Labelling differences between model train and model test data can lead to pragmatic failure of the metric evaluation of the model which presents as misleading values (shown as dashed arrow).}
\label{fig:triad_pragmatic}
\end{figure}

\subsection{Contribution}

We introduce a metric and labelling method which together provide robustness to granularity mismatches between ground truth and predictions, as well as an interpretable error which decomposes into distinct page parsing error modes. The framework is designed for natively 2D media, works with both Bounding Box and Mask based models, providing higher pragmatic competence across labelling schemas. The four main contributions are as follows:

\begin{enumerate}
    \item \textbf{The Missing, Overlap, Trespass and Excess (MOTe) error}: A decomposable evaluation metric for analysing model performance across different failure modes. It provides granularity robustness by allowing multiple predictions to be assigned to a single ground truth region, and penalises ground truth missed by the predictions, as well as predictions that would create parsing-induced character errors. 
    \item \textbf{Structural Semantic Unit (SSU)}: A relational labelling method that allows semantically related, but structurally distinct ground truth regions to act as a single object. This provides granularity robustness by allowing a single prediction to be assigned to multiple ground truth regions within the same SSU.
    \item \textbf{Relationship between parsing and character errors}: We derive idealised bounds for parsing-induced character error using the MOTe components, and show empirically that MOTe tracks these errors more closely than F1 or Dice.
    \item \textbf{MOTe library}: We release an Open Source Python library, \verb|mote_error|, for using this framework in analysis to support the community in Document Understanding tasks.
\end{enumerate}

In addition, we provide a structurally and semantically labelled update of the NCSE dataset \cite{bourne_reading_2025}, to facilitate scholars analysing models using MOTe.

Our discussion overlaps with ideas in ML and statistics such as dataset shift \cite{quinonero-candela_dataset_2009}, construct validity \cite{freiesleben_benchmarking_2025}, and label noise \cite{northcutt_pervasive_2021}: a metric is only informative to the extent that the data, labels, and metric together capture the construct we care about. Our use of Wittgensteinian ``meaning as use" and the language of pragmatic failure is intended to highlight this point from a complementary, semiotic perspective.

\section{Motivation}

Considering the importance of the IoU and the value it has provided machine vision practitioners it is important to provide a conceptual overview and intuitive introduction to the Structural-Semantic Unit (SSU), and the MOTe error.

\subsection{Motivating the MOTe}

The MOTe error has three main elements Missing, Overlap, Trespass; and a support metric Excess. With the exception of Excess the metrics are defined relative to the area of the ground truth, which unlike IoU based metrics, means any areas of prediction outside a ground truth do not count towards the final error. A qualitative definition of the metrics is given in \autoref{tab:metrics}, whilst the formal definition is given in \autoref{sec:definition}. The table shows the parsing-induced character error that occurs as a result of each MOTe component, Missing results in missing characters, Overlap creates duplicates, Trespass results in mis-ordered characters, and Excess has no impact as it is blank space. It is the fact that the MOTe error is relative to the area of the ground truth combined with the ability to assign many predictions to a single ground truth, that provides the granularity robustness as smaller valid predictions, (e.g. lines), can be aggregated to a ground truth at a higher level, (e.g. paragraphs).

\begin{table*}[h]
\centering
\begin{tabular}{|l|p{5cm}|p{4cm}|l|}
\hline
\textbf{Metric} & \textbf{Definition} & \textbf{Impact} & \textbf{PICE type}\\
\hline
Missing & How much of the ground truth is not covered by predictions & Penalises uncovered content & Missing\\
\hline
Overlap & How much of the ground truth has more than one prediction covering the same area & Penalises stacked predictions, impossible in 2D space & Duplicated\\
\hline
Trespass & How much of the ground truth is covered by a prediction that covers semantically or structurally distinct content & Penalises breaches of the semantic structure & Mis-ordered\\
\hline
Excess & How much area outside the ground truth is covered by predictions & Contextualises the core metrics & None\\
\hline
\end{tabular}
\caption{Qualitative definitions of the MOTe components and the parsing-induced character error (PICE) each is associated with.}
\label{tab:metrics}
\end{table*}

\begin{figure*}
    \centering
    \includegraphics[width=\linewidth]{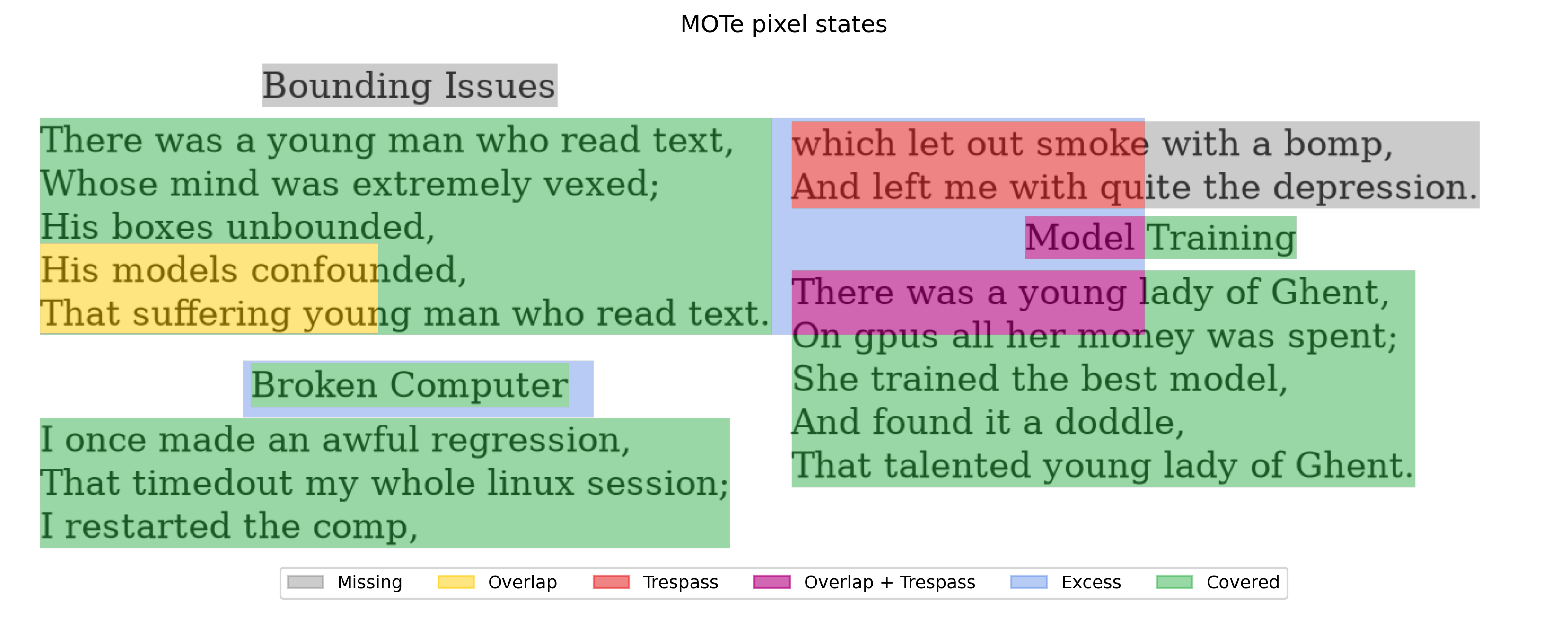}
    \caption{Example providing a visual representation of the MOTe error. The example shows how predictions and ground truth interact.}
    \label{fig:cot_components}
\end{figure*}

For a visual example of the elements of MOTe see Figure \ref{fig:cot_components} which shows the text of three limericks overlaid with  prediction masks. With only a single prediction covering the ground truth the mask is green, overlapping predictions are yellow, predictions that trespass into an adjacent limerick are in red, predictions that trespass and overlap are in purple, predictions that cover an area which has no corresponding ground truth are shown in blue, finally areas which do have ground truth but are not covered by a prediction are shown in grey. It is clear from the figure that Overlap will result in duplicate characters being extracted during OCR, whilst Trespass will result in characters being extracted out of reading order. These parsing-induced character errors, cause downstream issues both for evaluating OCR and analysing the resultant corpus and so should be directly penalised.

\subsection{The SSU}
\label{sec:motivation_ssu}

Typically DLA datasets are labelled using physical elements, such as lines and paragraphs, not the semantic elements which give the text actual meaning. As mentioned in Section \ref{sec:intro}, this becomes a problem when there is a large difference in granularity between the ground truth and prediction labelling as the IoU, F1, and mAP rapidly collapse to zero. The SSU resolves this by shifting the focus of the ground truth labels from the physical position of the text, to the meaning that the text conveys by grouping the labels of the physical text into common semantic and layout units. Ground truth regions that share the same Semantic and Structural units form a single SSU. This provides granularity robustness as larger predictions (e.g. paragraphs), are measured relative to a collection of smaller ground truth regions (e.g. lines), that have been aggregated together in the form of an SSU.

It can be seen that the SSU provides the complementary granularity robustness to the MOTe, providing the many-to-one-to-many mapping shown in Figure \ref{fig:ssu_pivot}, and ensuring that labelling and model granularity, does not impact the model-data-metric triad, and is no barrier to measuring parsing quality. 

\begin{figure*}
\centering
\begin{tikzpicture}[
    box/.style={
        draw=black!80, rounded corners=6pt, fill=white,
        minimum width=2.2cm, minimum height=0.9cm,
        font=\small, line width=0.8pt
    },
    ssubox/.style={
        draw=black!80, rounded corners=6pt, fill=black!5,
        minimum width=2.4cm, minimum height=1.6cm,
        font=\bfseries\small, line width=1.2pt, align=center
    },
    arr/.style={->, line width=0.8pt, >=Stealth, shorten >=3pt, shorten <=3pt}
]

% Left column: physical labels (3)
\node[box] (l1) at (0, 2.4)  {GT region 1};
\node[box] (l2) at (0, 1.2)  {GT region 2};
\node[box] (l3) at (0, 0)    {GT region 3};

% Center: SSU pivot
\node[ssubox] (ssu) at (3.2, 1.2) {SSU};

% Right column: predictions (4) -- deliberately asymmetric count
\node[box] (p1) at (6.4, 3.0)  {Pred A};
\node[box] (p2) at (6.4, 1.8)  {Pred B};
\node[box] (p3) at (6.4, 0.6)  {Pred C};
\node[box] (p4) at (6.4, -0.6) {Pred D};

% Both sides point INTO the SSU -- it is the target of two independent mappings
\draw[arr] (l1) -- (ssu);
\draw[arr] (l2) -- (ssu);
\draw[arr] (l3) -- (ssu);

\draw[arr] (p1) -- (ssu);
\draw[arr] (p2) -- (ssu);
\draw[arr] (p3) -- (ssu);
\draw[arr] (p4) -- (ssu);

\node[font=\footnotesize, align=center] at (0, -1.0)   {many-to-one\\(labelling)};
\node[font=\footnotesize, align=center] at (6.4, -1.4) {many-to-one\\(assignment)};

\end{tikzpicture}
\caption{The SSU as the common mapping of two independent many-to-one mappings. Ground-truth region labels (left) are grouped into a single SSU, and any number of predictions (right) are independently assigned to that same SSU. The two sides need not match in count, which is the source of the framework's granularity robustness.}
\label{fig:ssu_pivot}
\end{figure*}

Given the issues around meaning as use and the subjectivity of labelling discussed in Section \ref{sec:intro}, we only provide a general guide to defining class, structural unit, and semantic unit, as such definitions are up to the practitioner. In this way SSU are similar to classes because they are both project and use case specific. By making these definitions, the practitioner moves the data from being defined by an arbitrary spatial geometry to an explicitly defined semantic one. Algorithm \ref{alg:SSU} provides guidance to identifying SSU. The Algorithm shows that for two regions to be part of the same SSU they must be adjacent in reading order, of the same class, within the same structural unit, and be part of the same semantic unit. 

As an example of the SSU structure we can use the limericks in Figure \ref{fig:cot_components}. Each limerick, including its title, are their own Semantic Unit, whilst the columns create two structural units. The first and last limericks are broken into two SSUs each, with the title and text being separate (different pseudo-structures). The 2nd limerick is broken into three SSUs as the text is split across both columns. Supplementary Material Section 3 provides a simple example of granularity robustness in both directions. For additional practical guidance we also provide, the labelled NCSE dataset, and a labelling function for the PAGE labelling schema in the Python library.

\begin{algorithm}
\caption{SSU Region Assignment}
\label{alg:SSU}
\begin{algorithmic}[1]
\State Region $B$ belongs to the same SSU as Region $A$ if:
\State \quad $\text{same\_class}(A, B)$
\State \quad $\textbf{AND} \; \text{same\_structural\_unit}(A, B)$
\State \quad $\textbf{AND}  \; \text{same\_semantic\_unit}(A, B)$
\State \quad $\textbf{AND}  \; A \text{ and } B \text{ are adjacent}$
\end{algorithmic}
\end{algorithm}

\section{Defining MOTe error}
\label{sec:definition}

Having motivated MOTe error we now provide a formal definition of each component.
First to understand the mathematical underpinning of the MOTe consider a document image represented by the pixel matrix $M$ of dimensions $(H\times W)$, each SSU can be represented as $M^\mathrm{S}_i \in \{0,1\}$. As SSU are non-overlapping the binary mask representing the area covered by all $m$ SSU is $M^\mathrm{S} = \sum^m_i M^\mathrm{S}_i $, which is the composition of all sub-masks. Each prediction can be represented by the binary mask $M^\mathrm{p}_j \in \{0,1\}$. The composed sum of the masks is $M^\mathrm{p} \in \{0,1,..., \tilde{n}\}$ which is an integer mask bounded by 0 and $\tilde{n}$ the total number of predictions. To know the total area covered by the predictions we binarise $M^\mathrm{p}$ such that $M^\mathrm{p,b} \in \{0,1\} = M^\mathrm{p}>0$. The total area of the SSU can be found using $A^\mathrm{S} = \sum_{x,y}M^\mathrm{S}$.

\subsection{Assigning predictions to an SSU}
\label{sec:assign_ssu}
Similar to the IoU a prediction needs to be assigned to a ground truth in order to be evaluated. When using the MOTe a prediction is assigned to the appropriate Structural-Semantic Unit (SSU) based on whichever SSU makes up most of the prediction's area. The MOTe permits multiple predictions to be assigned to the same SSU. Predictions which do not overlap any SSU only count towards the support metric Excess. 

A prediction is assigned to an SSU using the following

\begin{equation}
    i^* = argmax_i \;\sum_{x,y}M^\mathrm{p}_j \odot M^\mathrm{S}_i
\end{equation}

which is the SSU $i$ ($M^\mathrm{S}_i$) that maximises the sum of the element wise product ($\odot$) with the binary mask of prediction $j$ ($M^\mathrm{p}_j$). In the case that there is no overlap between $M^\mathrm{p}_j$ and any SSU the prediction is not assigned an SSU. As such the $\tilde{n}$ total predictions are reduced to $n$ assigned predictions. The unassigned $\tilde{n}-n $ predictions do not affect the resultant text, as they contain no semantic content, they are also not included in the normalisation process discussed below. In the case that the prediction has equal coverage between more than one SSU, assignment is arbitrarily given to the SSU with the lowest index, reflecting reading order, ensuring a deterministic tie-breaker, as SSU index is a result of the page layout.

\subsection{Missing}

Missing error is defined through Coverage, which describes the relationship between the predictions and the ground truth, it is found by

\begin{equation}
\mathcal{C} = \frac{\sum_{x,y} M^\mathrm{S} \odot M^\mathrm{p,b} }{A^\mathrm{S}}
\label{eq:coverage}
\end{equation}

Thus making Coverage the normalised area of the intersection between the prediction and SSU masks, where $0 \leq \mathcal{C} \leq 1$, with $1$ being perfect coverage and 0 being no coverage. The Missing error is the complement of Coverage

\begin{equation}
\mathcal{M} = 1 - \mathcal{C}
\label{eq:missing}
\end{equation}

where $0 \leq \mathcal{M} \leq 1$, with $0$ being nothing missing and 1 being no coverage. This is a significant difference with the IoU as if an SSU has no assigned prediction the overall Missing rises removing the need for the `False negative'.

\subsection{Overlap}
The Overlap is concerned with the relationship between predictions. Overlap results in repeated phrases and sentences, and impacts the overall semantic coherence. Overlap is found by

\begin{equation}
    \mathcal{O} = \frac{\sum_{x,y}M^\mathrm{S} \odot (M^{\mathrm{p}} - M^{\mathrm{p,b}}) }{A^\mathrm{S}}
\end{equation}

Which is the duplicated coverage as a result of intersecting predictions normalised by the area of the ground truth.

\subsection{Trespass}

Trespass is concerned with the relationship between each prediction and the SSUs to which it is not assigned. Such Trespass causes the merging of semantically unrelated text producing word order errors during the OCR phase. The trespass of an individual predicted box is 

\begin{equation}
t_j= \frac{\sum_{x,y} \;M^{\mathrm{S}}_{\backslash i} \odot M^\mathrm{p}_j }{A^\mathrm{S}}    
\end{equation}

Where $M^{\mathrm{S}}_{\backslash i} = \sum_{k=1, k \neq i}^{m} M^{\mathrm{S}}_k$ is the ground truth mask excluding the region assigned to prediction $j$; the numerator $\sum_{x,y}(M^{\mathrm{S}}_{\backslash i} \odot M^\mathrm{p}_j) $ is the area of the intersection of predicted box $j$ and all SSUs that are not in $j$'s assigned ground truth SSU $i$. The trespass of an individual box is normalised across the area of all SSUs.

As such the Trespass score for the entire image is 

\begin{equation}
\mathcal{T}= \sum_j t_j=  \frac{ \sum_j\sum_{x,y} M^{\mathrm{S}}_{\backslash i} \odot M^\mathrm{p}_j}{A^\mathrm{S}}    
\end{equation}

Which is the sum of the trespass scores for all predicted boxes.

\subsection{The support metric: Excess}

Excess measures the total predicted area outside the bounds of any SSU. Excess shows how well the predictions fit the SSUs and any tendency to predict purely blank space. However when there are significant differences in granularity such as annotation at character level and prediction at paragraph level, the inter annotation spacing will inflate the excess values limiting its diagnostic power. As such we treat Excess as a support metric and refer to it using a lower case $e$. To calculate the Excess we find the negative space $\mathcal{N}$ not covered by the SSU, using 

\begin{equation}
    \mathcal{N} = J -M^\mathrm{S}
\end{equation}

Where $J$ is a matrix of ones the same dimension as $M$
The Excess is defined at the level of individual predictions as

\begin{equation}
    E_j = \frac{\sum_{x,y} \mathcal{N} \odot M^\mathrm{p}_j}{A^\mathcal{N}}
\end{equation}

Which is the intersection of the white space and the $j$th prediction normalised by the total area of the white space $A^\mathcal{N}$. The Excess at image level is shown as

\begin{equation}
    E = \frac{\sum_{x,y} \mathcal{N} \odot M^\mathrm{p,b}}{A^\mathcal{N}}
\end{equation}

Which is the same as for an individual prediction except using the binary mask of all predictions. The binary mask is used as the area outside SSUs have no semantic content and so does not impact downstream activities. Similar to Missing, Excess is naturally bounded between 0 and 1. It is maximised if the predictions cover the entire image. Low Missing with high Excess indicates that information is being effectively identified but that the predictions extend into the page margins and gutters, which may or may not be a problem. High Missing and High Excess indicate that the model is poorly trained and failing to find the regions of interest.

\subsection{Overall MOTe error}

The overall MOTe error is calculated as 
\begin{equation}
    \mathrm{MOTe} = \mathcal{M} + \mathcal{O} + \mathcal{T} 
\end{equation}

That is the sum of Missing, Overlap, and Trespass. Although this framing doesn't use Excess, it is still called MOTe for simplicity. 
The additive structure means that 0 is a perfectly parsed result, 1 occurs when there are no predictions, and a value greater than 1 means that the sum of Trespass and Overlap is larger than the total Coverage of the predictions. The theoretical upper bound is $2n-1$ where $n$ is the total number of predictions, whilst the practically attainable upper bound is $2n - 1 - n\frac{A^\mathrm{S}_\mathrm{max}}{A^\mathrm{S}}$ where $A^\mathrm{S}_\mathrm{max}$ is the largest SSU, derivation is provided in Supplementary Material 1. A single bounding box across the entire image would produce scores of Missing=0 Excess=1, Overlap=0, and a Trespass of $\mathcal{T} =1- \frac{A^\mathrm{S}_\mathrm{max}}{A^\mathrm{S}}$.

Due to its mask based construction, MOTe operates on any set of per-instance masks and so applies to object detection, instance segmentation and panoptic segmentation. A side effect of this is that when evaluating panoptic segmentation models Overlap is guaranteed to be zero by construction. 

Whilst the components of MOTe error could be weighted the basic structure is additive because each element represents the same spatial area, making equal weights the null hypothesis. However, in the python library the weights can be customised.

Simple worked examples of the MOTe building on the limerick example shown in Figure \ref{fig:cot_components} are provided in Supplementary Material Section 3.

\subsection{Multi-class analysis}

The MOTe error has so far been described in a class agnostic way, however, it is common to analyse document layout models against documents that contain multiple classes. Although multi-class analysis is not performed in this paper we include the framework for such an analysis both here and in the python library. As such we extend the definition to $K$ classes, where $\mathcal{M}_k$, $\mathcal{O}_k$, and $\mathcal{T}_k$ denote the Missing, Overlap, and Trespass for predictions of class $k$ respectively.

Predictions and SSUs each carry a class label $k \in \{1, \dots, K\}$. We write $A^{\mathrm{S}}_k = \sum_{x,y} M^{\mathrm{S}}_k$ for the total ground truth area of class $k$, where $A^{\mathrm{S}} = \sum_k A^{\mathrm{S}}_k$. 

\label{sect:cotef1}
We should note that Equation \ref{eq:coverage} is functionally the pixel level Recall of the prediction where $\sum_{x,y} M^\mathrm{S} \odot M^\mathrm{p,b}$ is the sum of the true positives. We can create the Precision score by dividing the sum of the true positives by the total predicted area. By incorporating class into the equations we find 

\begin{equation}
    \mathrm{Recall}_k = \mathcal{C}_k = 1 - \mathcal{M}_k = \frac{\sum_{x,y} M^\mathrm{S}_k \odot M^\mathrm{p,b}_k}{\sum_{x,y} M^\mathrm{S}_k}
\end{equation}

\begin{equation}
  \mathrm{Precision}_k= \frac{\sum_{x,y} M^\mathrm{S}_k \odot M^\mathrm{p,b}_k}{\sum_{x,y} M^\mathrm{p,b}_k}
\end{equation}

We can then calculate the F1 score being the harmonic mean of precision and recall

\begin{equation}
    F_1 = 2 \cdot \frac{\text{Precision} \cdot \text{Recall}}{\text{Precision} + \text{Recall}}
    \label{eq:f1}
\end{equation}

as such the pixelwise F1 score for class $k$ can be shown to be

\begin{equation}
    F_{1,k}= 2 \frac{\sum_{x,y} M^\mathrm{S}_k \odot M^\mathrm{p,b}_k}{\sum_{x,y} M^\mathrm{S}_k + \sum_{x,y} M^\mathrm{p,b}_k}
    \label{eq:dice_score}
\end{equation}

which is the Dice coefficient or pixel-wise F1 positioning the MOTe error's pixel-level approach with segmentation-style evaluators. The relationship with image segmentation approaches and natural scenes is shown using examples in Supplementary Material Section 5. The MOTe F1 can be aggregated across classes using either micro or macro averaging approaches. Using this pixel based $F_{1}$ means the MOTe natively produces scalar based evaluation of class detection quality in a way that the core MOTe components alone do not, although it is relative to Coverage, discarding information on Overlap through binarisation, and folding cross-class (but not within class) Trespass and Excess into the precision term. This pixel based $F_{1}$ also maintains the granularity robustness that the IoU-based F1 lacks. 

Additional diagnostics such as confusion matrices are shown in Supplementary Material Section 2

\section{The MOTe, F1, and the CER}
\label{sect:cote_f1_cer}

Given that a large amount of Document Understanding relates to the extraction of text from images of pages, it is of value that page parsing metrics relate to the resultant quality of text. This link with character errors is because parsing-induced character errors create a structurally irreducible error floor which cannot be resolved by an OCR engine. As discussed in Section \ref{sec:intro} the F1 score is the dominant metric for evaluating DLA. The F1 defined in Equation \ref{eq:f1}, and the Precision and Recall terms are

\begin{equation}
    \text{Precision} = \frac{\text{TP}}{\text{TP} + \text{FP}} = P,
\quad
\text{Recall} = \frac{\text{TP}}{\text{TP} + \text{FN}} =R
\end{equation}

where TP is True Positive, FP is False Positive, and FN is False Negative. These along with TN, True Negative, are the 4 possible truth states. We can map these truth states to the MOTe as shown in table \ref{tab:map_to_cote}. The table shows that the F1 cannot express Trespass, modifies the score based on areas that are typically empty (excess), and due to the one-to-one matching cannot differentiate between Excess and Overlap. We describe the F1 as having low semantic competence due to it's weak ability to interpret the structural semantics of the document, as whilst it is able to measure whether things are there or not, it cannot appreciate the relationship between them.

\begin{table}
    \centering
    \begin{tabular}{|l|l|}
    \hline
        \textbf{Type}   & \textbf{MOTe}\\
        \hline
        TP & $\downarrow$ Missing\\
        FN & $\uparrow$ Missing\\
        TN & no impact\\
        FP & $\uparrow$ Excess or Overlap\\
\hline
    \end{tabular}
    \caption{How different truth types relates to the MOTe}
    \label{tab:map_to_cote}
\end{table}

The following two sections show how the differences in the structure of the MOTe and F1 impact how they relate to the CER.

\subsection{Finding the bounds of the CER in relation to the MOTe}
\label{sect:derive_cote_cer_bounds}

\begin{figure*}
    \centering
    \includegraphics[width = \linewidth]{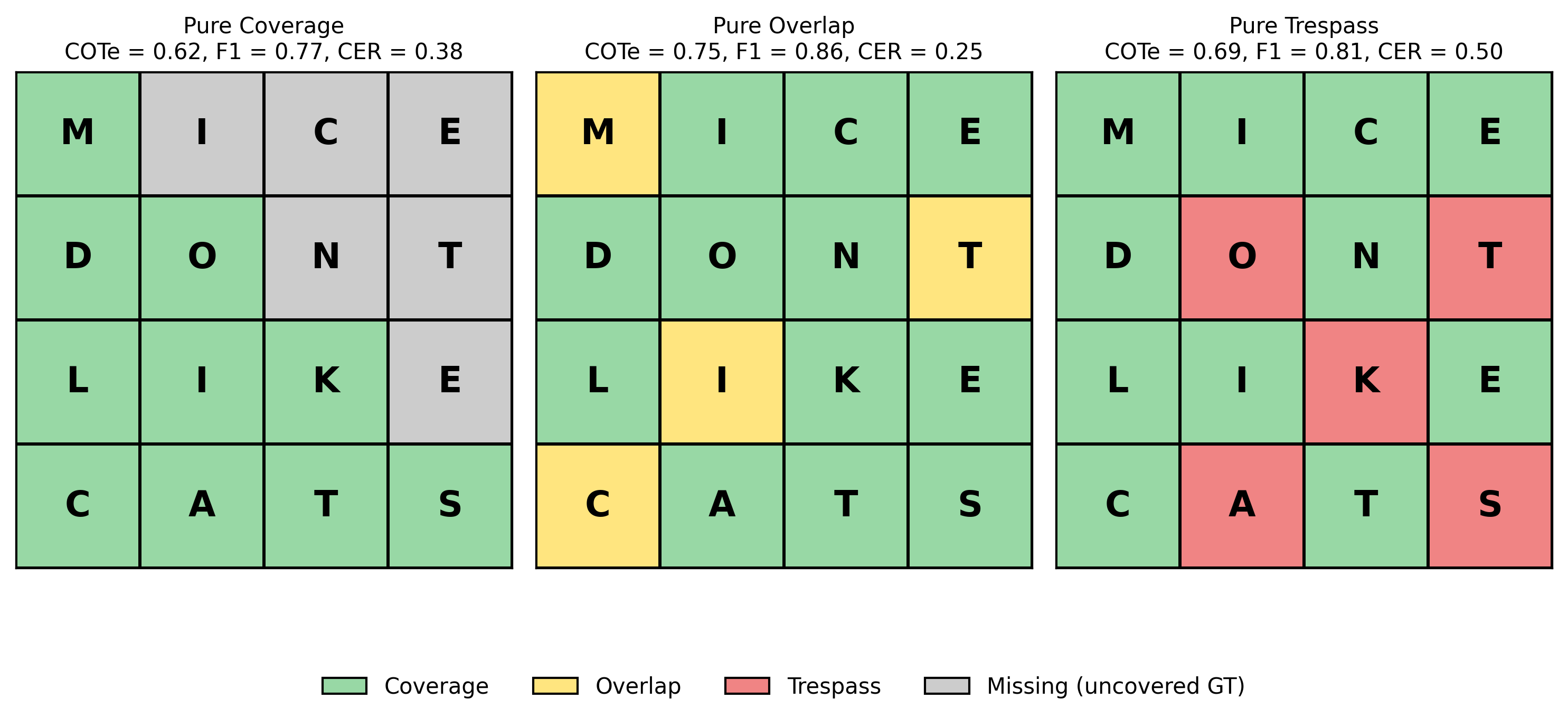}
    \caption{Three different types of pure error.}
    \label{fig:pure_error}
\end{figure*}

Because the MOTe penalises infringements on the semantic structure it provides ability in tracking the CER. What is more the lower and upper bounds of the CER can be found in terms of the error components of the MOTe. Figure \ref{fig:pure_error}, shows a simple example where each character is a single SSU being read left to right, each panel only includes pure errors, that is only one type of error is present. In the first panel, there are Missing errors, in panel 2 Overlap errors, in panel 3 Trespass errors.

The parsing errors shown in the figure result in reading order errors even if the OCR is perfect. These errors are shown in Table \ref{tab:pure_error}. Letters that have been deleted are shown in blue, letters that have been inserted are shown in orange. There are two Trespass lines one with errors and one without. This is because the Trespass could be caused by the bounding box extending from the top down creating an error in reading order, or from the left to the right, meaning no reading order errors are induced. 

\begin{table}[t]
\centering
\caption{Reading order errors due to parsing. Blue letters have been deleted, Orange letters have been inserted}
\label{tab:pure_error}
\begin{tabular}{|l|l|}
\hline
\textbf{Error Type} & \textbf{Reading Order} \\
\hline
Missing       & M\greyl{ICE} DO\greyl{NT} LIK\greyl{E} CATS \\
Overlap       & \redl{M}MICE DON\redl{T}T L\redl{I}IKE \redl{C}CATS \\
Top Trespass  & MI\redl{O}CE\redl{T} D\greyl{O}N\redl{K}\greyl{T} LI\redl{A}\greyl{K}E\redl{S} C\greyl{A}T\greyl{S} \\
Left Trespass & MICE DONT LIKE CATS \\
\hline
\end{tabular}
\end{table}

What the figure shows is that Missing errors result in a deletion per character uncovered, Overlap errors result in a single insertion per character overlapped, and Trespass may require no changes or up to 1 deletion and 1 insertion. The CER induced by Trespass shown in Figure \ref{fig:pure_error} is 0.5, when the theoretical error from 5 trespasses is $10/16=0.63$, this difference is because adjacent deletions and insertions can merge into a single substitution reducing the total required change count and lowering the calculated CER. As a result, the CER for each pure error type can be shown as

\begin{equation}
    \mathrm{CER}_\mathcal{M} = \mathcal{M}
    \label{eq:pure coverage}
\end{equation}

\begin{equation}
    \mathrm{CER}_\mathcal{O} = \mathcal{O}
\end{equation}

\begin{equation}
    \mathrm{CER}_\mathcal{T} \in [0, \mathrm{min}(1, 2\mathcal{T})]
    \label{eq:trespass_cer_limit}
\end{equation}

Where Trespass is a range that is capped at 1 as in the worst case where all characters are covered by a single prediction and are out of place, there would need to be $l$ substitutions for a string of length $l$ resulting in a maximum CER of 1. Trespass is shown as being an interval due to Trespass not necessarily causing parsing-induced character errors and, as mentioned earlier, the possibility of merging adjacent errors.

As the MOTe is additive and in the worst case the errors are entirely non-overlapping it means that we can provide idealised bounds given two assumptions (A1) The document filled with characters of uniform density. (A2) The reading order within SSU is recoverable (e.g. lines are not split into two separate regions). Given these assumptions the CER should fall between

\begin{equation}
    \mathrm{max}(\mathcal{M}, \mathcal{O})\leq \mathrm{CER} \leq \mathcal{M} + \mathcal{O} + 2\mathcal{T}
\end{equation}

The lower bound is composed of Missing and Overlap because, under pure error, errors of these types are guaranteed to produce CER. The Trespass is not capped at 1 as in Equation \ref{eq:trespass_cer_limit} because Trespass can stack when Overlap is present meaning CER related to trespass can extend beyond 1. The lower bound is exact in the case of pure Overlap or Missing errors, interaction between error types, and violations of the assumptions can make these bounds soft in practice. The formula also suggests that $\mathrm{CER} \lesssim \mathrm{MOTe}$ will work when Trespass is small relative to total MOTe error and will understate the CER by at most $\mathcal{T}$. 

Importantly, given assumption A2 is true, when the page is perfectly parsed and MOTe$=0$ we find that CER=0, also that when there are no predictions MOTe$=1$ and CER=1. However it does not hold that when CER=1, that MOTe=1.

\subsection{Bounds on MOTe F1/Dice}

Because the MOTe F1 or Dice score shown in Equation \ref{eq:dice_score} is area based a lower bound can also be found but only under the following two assumptions (A1) Densely packed uniformly distributed characters, (A2) Precision = 1. In that case Recall is simply Coverage, $1-\mathcal{M}$, so

\begin{equation}
    D = 2\frac{1-\mathcal{M}}{2-\mathcal{M}}
\end{equation}

Where $D$ is the Dice score, rearranging gives

\begin{equation}
    \frac{2(1-D)}{2-D} = \mathcal{M}
\end{equation}

as we know that in pure Missing errors $\mathrm{CER} = \mathcal{M}$ we can substitute

\begin{equation}
    \mathrm{CER}\geq \frac{2(1-D)}{2-D}
\end{equation}

Similar to the MOTe the bounds are idealised and so will be soft in practice.

\subsection{F1 and the CER}
\label{sect:f1_cer}

\begin{figure*}
    \centering
    \includegraphics[width=\linewidth]{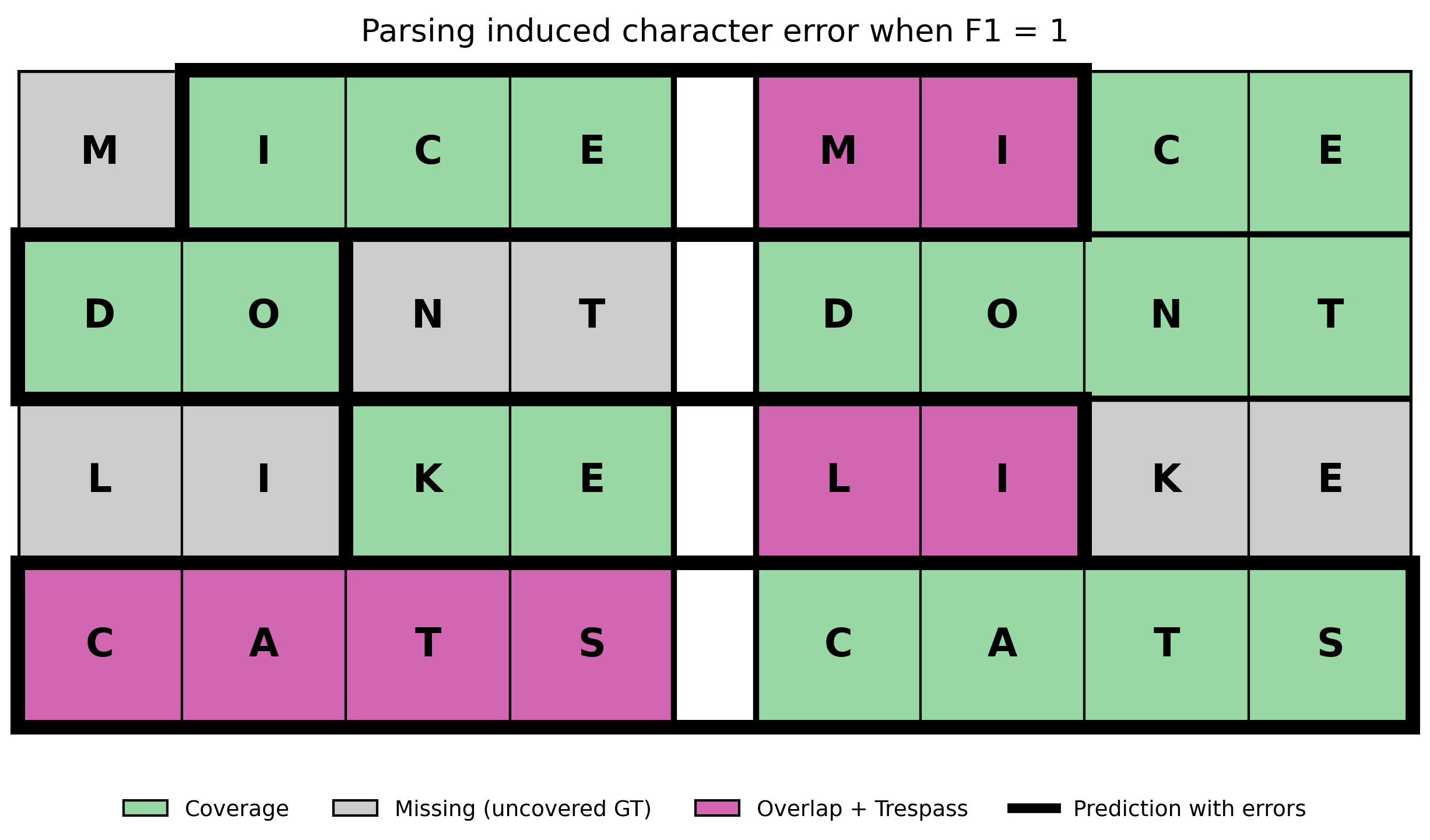}
    \caption{In this two column case the F1@50 produces a perfect score of 1, across the whole image and at line level, despite creating parsing-induced character errors in 6 of the 8 lines.}
    \label{fig:f1_failure_example}
\end{figure*}

As the F1 has no way of expressing Trespass and no clear way of distinguishing Overlap from Excess, it is clear that the metric is inherently less expressive than the MOTe. However, as the F1 only checks if predictions clear the IoU threshold, the F1 score does not tell us how well the regions were captured only that they were above a certain quality threshold. This complicates the interpretation of parsing quality even when the F1=1, as shown in Figure \ref{fig:f1_failure_example}. The Figure shows four obviously wrong configurations of prediction which induce character errors. However, at an IoU threshold of 0.5, each two column row, as an independent image, would score F1=1, as would the combined image. In contrast the MOTe would have highlighted parsing issues, and implicit parsing-induced character errors, as shown in Table \ref{tab:f1_failure_example}.

\begin{table}[]
    \centering
    \caption{Measured error by line}
    \label{tab:f1_failure_example}
    \begin{tabular}{lrrr}
    \toprule
     & F1@50 & MOTe & CER \\
    \midrule
    MICE & 1.00 & 0.62 & 0.38 \\
    DONT & 1.00 & 0.25 & 0.25 \\
    LIKE & 1.00 & 1.00 & 0.50 \\
    CATS & 1.00 & 1.00 & 0.50 \\
    \bottomrule
    \end{tabular}
\end{table}

This error mode can be generalised to any set where predictions of length $j+f$ are assigned to ground truth $k$ having an intersection of $j$, but extend into a non-reading order adjacent region by $f$, as such

\begin{equation}
    \mathrm{IoU} = \frac{j}{k + f}, \qquad \mathrm{CER} = \frac{(k - j) + f}{k}
\end{equation}

This means that for any IoU threshold less than 1, any true positive prediction can have a parsing-induced character error rate of

\begin{equation}
    \mathrm{CER}\leq 1-t
\end{equation}

if the prediction is smaller than the ground truth region and does not trespass. Alternatively, if the prediction is larger than the ground truth region and does trespass, the parsing-induced character error rate is

\begin{equation}
    \mathrm{CER}\leq \frac{1}{t}-1
\end{equation}

This approach can be extended to image level such that for a uniformly dense document with adjacent non-sequential text blocks the CER range for F1=1 at IoU=t is

\begin{equation}
 \mathrm{CER} \in \left[ 0, \frac{1}{t}-1 \right]    
\end{equation}

\begin{figure*}
    \centering
    \includegraphics[width=\linewidth]{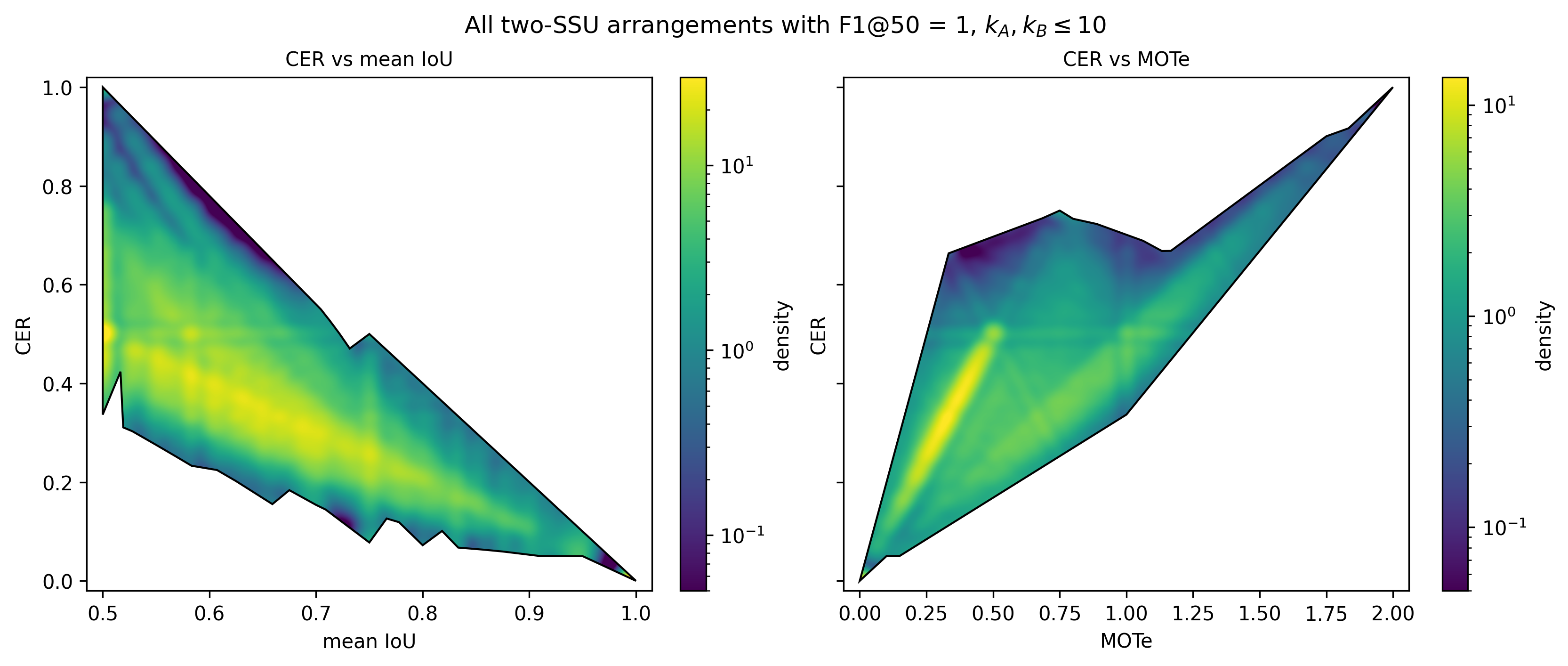}
    \caption{All combinations of the two SSU example where F1@50=1. The Figure shows that F1=1 does not guarantee no CER error whilst MOTe=0 does.}
    \label{fig:f1_pair_arrangements_cer}
\end{figure*}

We apply every combination of error for the case for values of $k$ up to 10 where F1@50=1, and plot the results in Figure \ref{fig:f1_pair_arrangements_cer}. Due to the quantity of points (41,743) we visualise using a density map clipped to the real boundaries of the underlying data. The Figure shows the mean IoU plotted against CER and visualises how F1=1 is no guarantee of low CER error. The irregular edges are an artifact of stopping at k=10.

\section{Method}

Having motivated the SSU and MOTe, defined them, and shown the theoretical link with CER, we now introduce the experiments used to validate them and explain their behaviours. We first introduce the datasets we will be using, we will then introduce a test of granularity robustness, and a test of correlation with CER. Finally, we will introduce the approach to model comparison.

In the experiments, we compare the MOTe, against the IoU, COCO mAP, and COCO style F1 at IoU$\geq 0.5$ (F1@50). As documents can include a large amount of bounding boxes we lift the matching cap on the COCO metrics from 100 to 500. In addition, as the paper flattens the data to a single class, to allow the models to be evaluated without fine-tuning, class mAP becomes simple Average Precision (AP). We have not used Panoptic Quality (PQ) \cite{kirillov_panoptic_2019} in this paper as it contains the F1 as a term so will face the same limit. However, for completeness, in Supplementary Material Section 5, we provide a brief example comparing MOTe and PQ when applied to natural scenes.

\subsection{Data}
\label{sect:data}

We will use three types of data, a publicly available data set in which we can infer character position, three publicly available DLA datasets which do not have detailed character information, and a highly controlled generated dataset we construct such that we know the position of every character.

\subsubsection{DLA datasets}
This paper will use four publicly available datasets to show the behaviour of the MOTe. We will use the NCSEV2 dataset \cite{bourne_ncse_2025}, the HNLA2013 dataset \cite{antonacopoulos_icdar_2013}, used in the 2013 ICDAR document layout competition, and The Spiritualist dataset \cite{nockels_spiritualist_2026}, all updated to include SSU. In addition, the popular layout benchmark DocLayNet \cite{pfitzmann_doclaynet_2022} which doesn't have SSU but provides opportunity to compare the models on a better understood dataset. The Spiritualist was transcribed using Transkribus, which provides both line-level and word-level segmentation. We use the line-level regions as the layout ground truth, and infer character positions from the word-level boxes using the method of \cite{bourne_character_2026}.

\begin{table*}[h!]
\centering
\caption{Details on the datasets used for evaluation, HNLA2013 and DocLayNet data are for their test sets, the NCSE dataset is only 31 pages in total}
\label{tab:dataset_info}
\begin{tabular}{|l|c|c|c|c|l|}
\hline
\textbf{Dataset} & \textbf{Image Type} & \textbf{Images} & \textbf{Regions} & \textbf{Region Type} & \textbf{SSU} \\
\hline
The Spiritualist & Newspapers & 49 & 425 & Polygon & Yes \\
NCSE & Newspapers & 31 & 358 & Bounding Box & Yes \\
HNLA2013 & Newspapers & 50 & 2668 & Polygon & Yes \\
DocLayNet & Multi Format & 4999 & 66531 & Bounding Box & No \\
\hline
\end{tabular}
\end{table*}

It is notable that none of the datasets used have a common labelling schema. The Spiritualist uses \cite{nockels_spiritualist_2026}, the NCSEv2 was only concerned with the accuracy of OCR not the layout, as such only text regions were tagged and simply mapped to the content. DocLayNet uses a more advanced, but still custom, labelling system that includes a collection of classes. HNLA2013 uses the PAGE system developed as a way to create an extensible standard for marking up pages of text. We use the test datasets of HNLA2013 and DocLayNet.

DocLayNet is designed to be large enough that models can be trained or fine-tuned on a training set then evaluated on the hold-out set, however, there is not enough data in the other three for training, and so models must rely on their own general pre-trained capabilities. For this paper we consider the case where custom training is not possible and simply compare the performance of models as is.

To handle different approaches to labelling we ignore label classes and simply look at region of interest comparability. Due to the rich structure of the PAGE format HNLA data was automatically labelled into SSU using the pre-defined page structure, and assuming semantic unit's are defined as the text between the labelled ``header" class, the functionality for this has been released as part of the python library. DocLayNet will not use SSU which means that the MOTe error will not have granularity robustness on this dataset.

\subsubsection{Controlled dataset}
\label{sect:controlled_dataset}
To measure parsing-induced character error we need to know exactly which characters are within a predicted region, to do this the position of every character and it's relative position within it's text group known. Such datasets are hard to obtain and expensive to make from scanned documents. To get around this we programmatically create a dataset of text documents. 

\begin{figure*}[htbp]
    \centering
    \begin{subfigure}[t]{0.32\textwidth}
        \includegraphics[width=\textwidth]{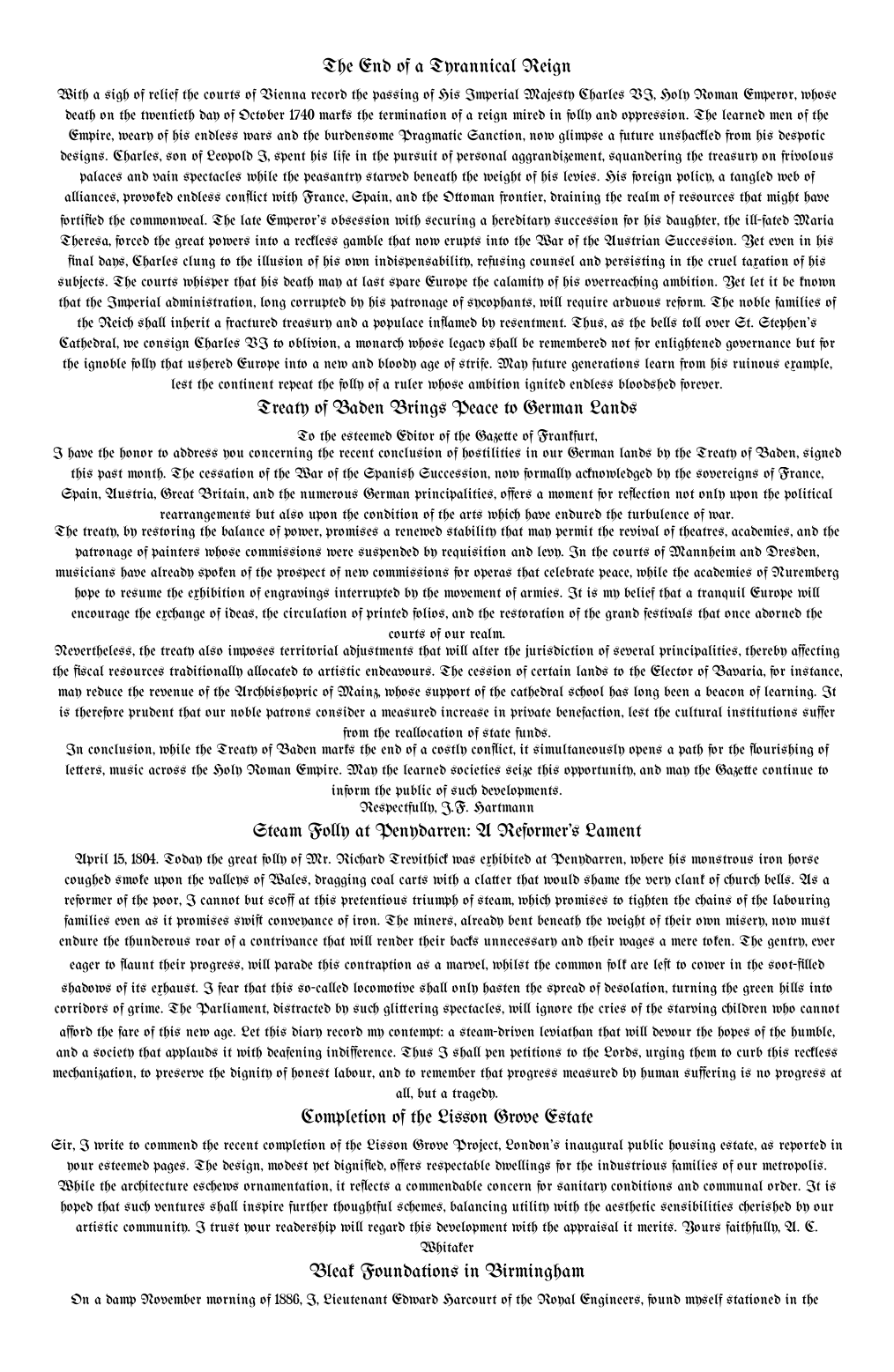}
        \caption{UnifrakturMaguntia centered 1 column}
        \label{fig:page1}
    \end{subfigure}
    \hfill
    \begin{subfigure}[t]{0.32\textwidth}
        \includegraphics[width=\textwidth]{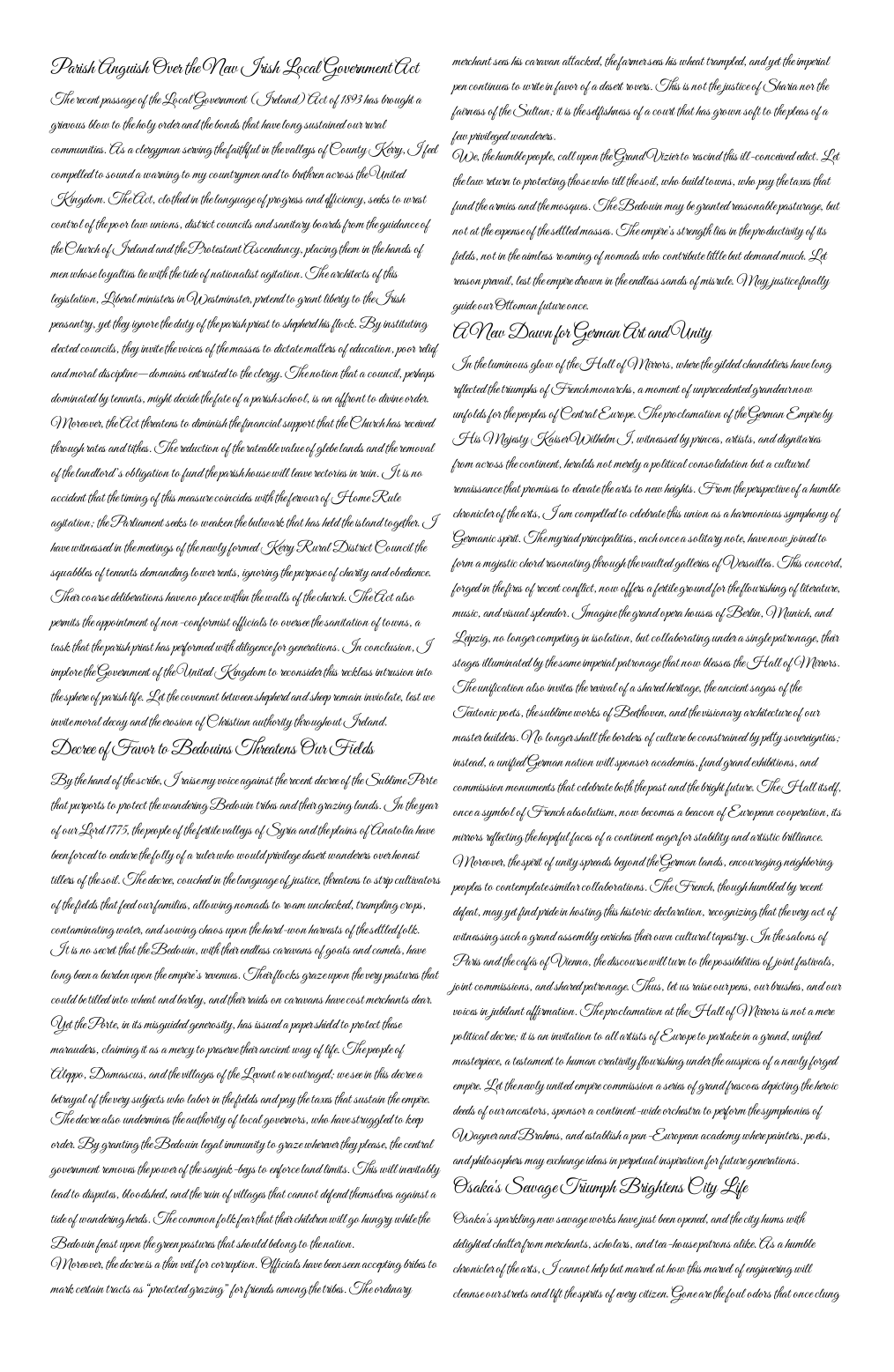}
        \caption{Great Vibes left aligned 2 columns}
        \label{fig:page3}
    \end{subfigure}
    \hfill
    \begin{subfigure}[t]{0.32\textwidth}
        \includegraphics[width=\textwidth]{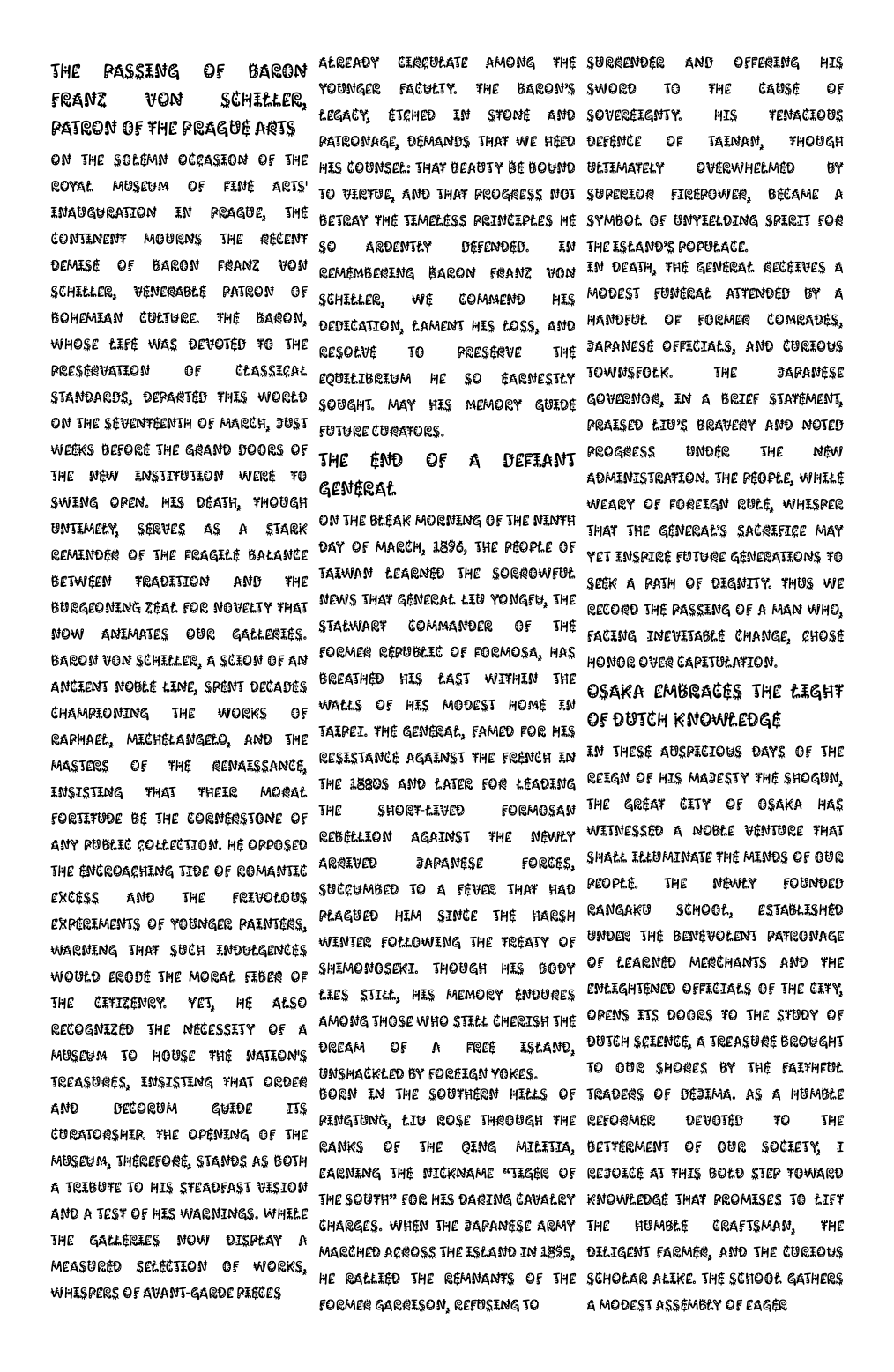}
        \caption{Kablammo justified 3 columns}
        \label{fig:page2}
    \end{subfigure}
    \caption{Example pages of text used for measuring granularity robustness.}
    \label{fig:comparison}
\end{figure*}

We do this by generating a large amount of synthetic text using an LLM to create articles based on historical events, as used in \cite{bourne_scrambled_2025}. We then use 5 fonts, three text alignment styles, and 1,2, or 3 columns, to create different layouts and formatting styles. Using these templates, we then generate 10 pages using each configuration, filled with randomly selected articles, this results in 450 pages overall. We hold page size constant at 280mm x 430mm. The font alignments will create different amounts of white-space. Figure \ref{fig:comparison} shows three example pages using different fonts, columns and alignments. For practical purposes the SSU only needs to be assigned at the largest granularity level in the experiment, which in this case is paragraph. The final dataset contains a total of 1595 articles, 418,544 words, and 2,152,985 characters excluding spaces.  
Detail on the configurations is shown in Table \ref{tab:synth_combos}.

\begin{table}[h!]
\centering
\caption{Parameters used in text generation process}
\label{tab:synth_combos}
\begin{tabular}{p{3.5cm}p{2cm}c}
\toprule
\textbf{Font} & \textbf{Layout} & \textbf{Columns} \\
\midrule
 Ewert, Great Vibes, Kablammo, Roboto, UnifrakturMaguntia  & Centered, Justified, Left aligned & 1, 2, 3 \\
\bottomrule
\end{tabular}
\end{table}

\subsection{Parsing models}

In this paper we will be using Doclayout-yolo \cite{zhao_doclayout-yolo_2024}, Heron \cite{livathinos_advanced_2025}, and PP-Doclayout \cite{sun_pp-doclayout_2025}. The models have been chosen as they are popular choices within Document Layout Analysis. Heron and PP-Doclayout-L  are both based on the same DETR architecture \cite{delestre_cmarkeadetr-layout-detection_2023}, while PPDoc-M and PPDoc-S are based on PicoDet \cite{yu_pp-picodet_2021}. Details of the models can be seen in Table \ref{tab:layout_models}, all models will be run using default parameters.  The models will be run on a 24GB NVIDIA L4. 

\begin{table*}[htbp]
\centering
\caption{Comparison of Document Layout Analysis Models}
\label{tab:layout_models}
\begin{tabular}{lccc}
\toprule
\textbf{Model} & \textbf{Type} & \textbf{Parameters} & \textbf{Framework} \\
\midrule
DocLayout-YOLO & CNN & 15.4M & PyTorch \\
Heron & DETR-based& 42.9M & PyTorch \\
PP-DocLayout-L & DETR-based  & 30.94M & PaddlePaddle \\
PP-DocLayout-M & PicoDet-M & 5.65M & PaddlePaddle \\
PP-DocLayout-S & PicoDet-S & 1.21M & PaddlePaddle \\

\bottomrule
\end{tabular}
\end{table*}

\subsection{Measuring granularity robustness}
\label{sec:meas_gran_robust}

We measure the impact of the many-to-one-to-many relationship afforded by the SSU using a controlled experiment, where predictions match the ground truth and all characters are extracted without error such that CER = 0. Using the controlled dataset described in Section \ref{sect:controlled_dataset}, we take every combination pair of character, word, line, and paragraph level ground truth, assign one member of the pair to be a prediction, and the other to be the ground truth, we will then measure the resulting MOTe and F1@50 scores. Because we are comparing ground truth bounding boxes on different scales the parsing-induced character error will be 0, however, we will expect to see the F1 collapse under granularity differences, whilst the MOTe error remains broadly stable for all variants, as the SSU is mapped to paragraph level. In this experiment F1 is computed directly, avoiding issues related to the COCO cap, particularly for character level analysis.

\subsection{Correlation with CER}

Although Section \ref{sect:cote_f1_cer} defines the relationship with CER in theoretical terms, we use the synthetic pages from Section \ref{sect:controlled_dataset} to empirically test the relationship between CER, MOTe, F1 and Dice. We do this using both parsing models and artificial predictions. To ensure that the synthetic data is comparable to real data we validate using the 49 page Spiritualist dataset.

For the parsing model predictions, we use each of the 5 parsing models to predict over the controlled dataset, we then calculate the MOTe, F1, and Dice scores. We then take advantage of having perfect character position information and reconstruct the page reading order based on the index of the first character in the predicted region. Subsequently we extract the exact text in each predicted region and generate page text under perfect OCR in the recovered reading order, where character order within the predicted regions is read top to bottom left to right as an OCR engine would. Finally we measure the Spearman correlation of parsing-induced character errors using CER against MOTe, F1, and Dice.

We perform ablations by combining parsing the ground truth of all pages with `predictions' based on the ground truth of all other pages stratified by number of columns. We then either leave with no noise, add noise from a page with the same number of columns, or add noise using a randomly selected page. This creates three sets of 202,050 page-prediction pairs, each with it's own type of error. No added noise, has low Missing and no overlap. Within prediction column count noise, has higher Missing and Trespass is mostly vertical if columns match and mostly horizontal if they don't. 
Random noise is the same as the within column count noise except Trespass direction is more randomised. For all noise types the Trespass on matching column counts will be dominated by vertical Trespass. Noise is added by randomly removing each prediction with page-level probability 0-0.5, the predictions box of a second random page are then added with page-level probability in the range 0-0.5. Because all content is random, the pages and bounding boxes are independent, controlling for the columns and to a lesser degree text alignment.

\subsection{Model and Dataset Comparison}

Having analysed MOTe's granularity robustness and correlation with CER, we will compare the DLA models and analyse the difference in both overall and decomposed performance.  

Whilst the MOTe error and F1, and mAP can both be applied at class level, for simplicity we will be treating all ground truth regions as a single class and only considering the quality of coverage of the SSUs in each page. This means that pseudo-structural elements such as titles, will be ignored and only structural elements such as columns, mastheads, etc will be considered. 

The goal of this analysis is not to see which model is better, but rather to gain deeper insight into their functioning, strengths, and potential failure modes, as well as the relationship between the metrics.

\section{Results}

The results are broken into three sections. First we measure the granularity robustness of the MOTe, then we explore the correlation with the CER, finally we provide examples of models applied to different datasets to help build intuition for the application of the MOTe, and its interpretation.

\subsection{Measuring Granularity robustness}
\label{sec:gran_robust}
Using the 450 generated pages of text, Tables \ref{tab:granularity_cote} and \ref{tab:granularity_f1} show how the MOTe and the F1 respond to different granularities in the prediction and ground truth when the image is perfectly parsed. As the upper triangle of the table is all 0, we see that the MOTe error is invariant to granularity when the ground truth is the same size as, or smaller than, the prediction. When predictions are smaller than the ground truth the MOTe error diverges from zero, incorrectly implying parsing error. The divergence from 0 of the MOTe is caused by empty space within the ground-truth boxes, examples of this are shown in Supplementary Material Section 3. However as shown in Table \ref{tab:granularity_f1}, the F1 which should be 1 throughout, collapses to 0 rapidly under granularity divergence. In the worst case the MOTe is off by 32 percentage points whilst the F1 is off by 100 percentage points, showing that compared to the F1 the MOTe is highly robust to granularity differences. Across all non-matching pairs the F1 diverged from the true value by 98 percentage points, in contrast the MOTe diverged by 9 percentage points. 

This shows the value of the many-to-one-to-many structure provided by the MOTe error. The standard deviation across the experiments is very low being max 0.04 for the MOTe and 0.06 for the F1, showing the results are quite stable. The mAP is even more sensitive to granularity differences than the F1 due to the higher IoU thresholds. Dice, produces a symmetric matrix where the values are the harmonic-mean of one minus the MOTe matrix and one minus it's transpose. This results in Dice not being invariant to granularity, but having more granularity robustness than MOTe when the prediction is smaller than the ground-truth. On average it performed slightly worse than MOTe having a mean divergence of 11 percentage points.

\begin{table}[htbp]
\centering
\caption{Mean MOTe error when one ground truth granularity is scored against another. All scores should be 0.}
\label{tab:granularity_cote}
\begin{tabular}{lcccc}
\toprule
\textbf{GT $\downarrow$ / Pred $\rightarrow$} & \textbf{Char} & \textbf{Word} & \textbf{Line} & \textbf{Para} \\
\midrule
Character & 0.00 & 0.00 & 0.00 & 0.00 \\
Word & 0.00 & 0.00 & 0.00 & 0.00 \\
Line & 0.10 & 0.10 & 0.00 & 0.00 \\
Paragraph & 0.32 & 0.32 & 0.24 & 0.00 \\
\bottomrule
\end{tabular}
\end{table}

\begin{table}[htbp]
\centering
\caption{Mean F1@50 when one ground truth granularity is scored against another. All scores should be 1.}
\label{tab:granularity_f1}
\begin{tabular}{lcccc}
\toprule
\textbf{GT $\downarrow$ / Pred $\rightarrow$} & \textbf{Char} & \textbf{Word} & \textbf{Line} & \textbf{Para} \\
\midrule
Character & 1.00 & 0.06 & 0.00 & 0.00 \\
Word & 0.06 & 1.00 & 0.01 & 0.00 \\
Line & 0.00 & 0.01 & 1.00 & 0.07 \\
Paragraph & 0.00 & 0.00 & 0.07 & 1.00 \\
\bottomrule
\end{tabular}
\end{table}

\subsection{Correlation with CER}
\label{sec:res_cor_cer}

\begin{figure*}
    \centering
    \includegraphics[width=\linewidth]{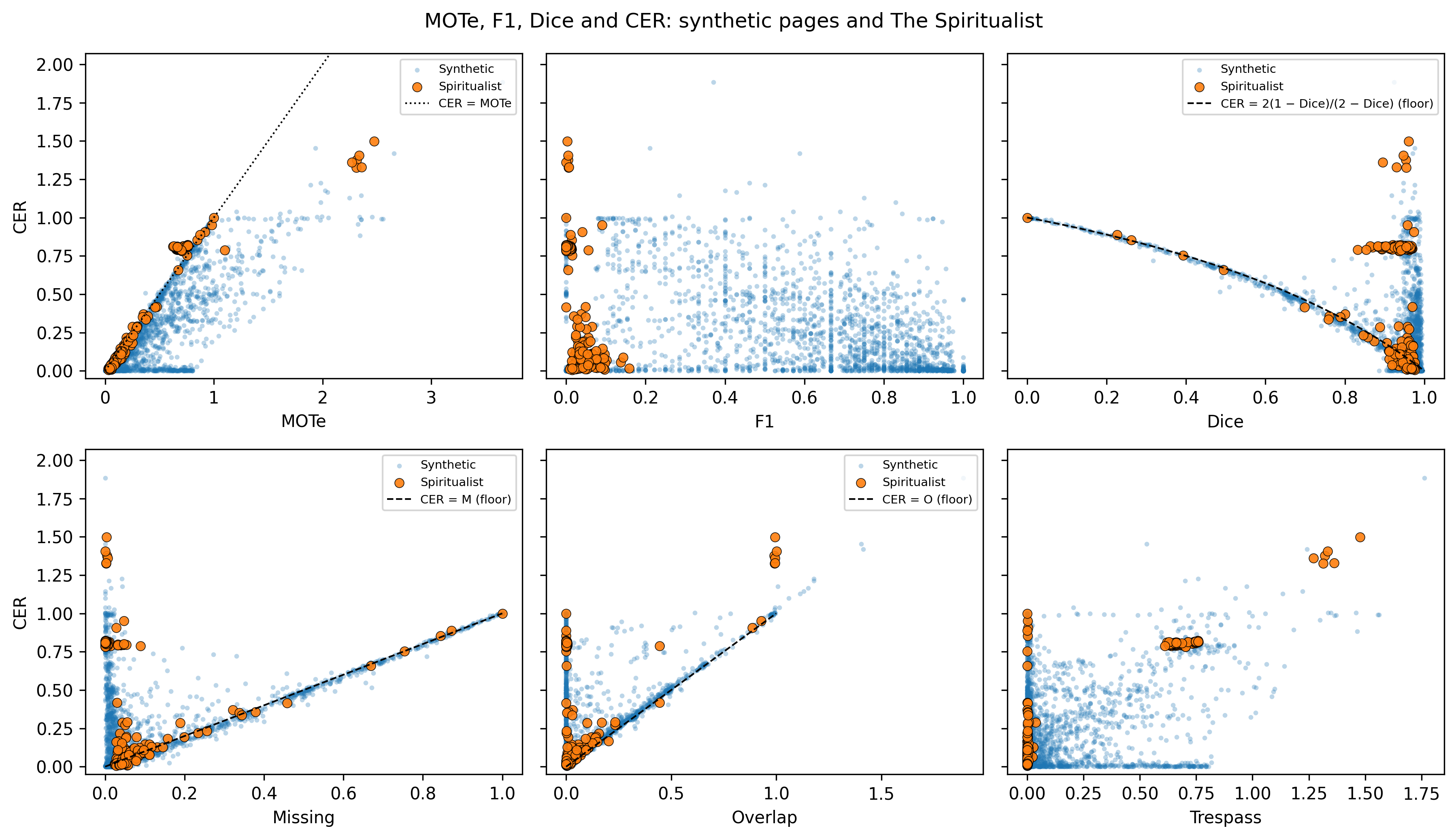}
    \caption{There are clear relationships between the MOTe, its decompositions and the CER. The derived CER bounds show high levels of adherence, even with noisy data.}
    \label{fig:spiritualist_cote_cer_scatter}
\end{figure*}

We compare the MOTe to the F1@50 and Dice across the 450 pages of generated text and 49 pages of The Spiritualist, as shown in Table \ref{tab:spiritualist_corr} and Figure \ref{fig:spiritualist_cote_cer_scatter}. The figure shows that the Synthetic data broadly follow the same pattern as The Spiritualist but covers more of the parameter space, likely due to more varied formatting. We also notice that PPDoc-M and PPDoc-S struggle with the task, making substantial Trespass errors on both the synthetic and The Spiritualist data (See Table \ref{tab:res_spiritualist}); therefore, we analyse with and without these two models. Table \ref{tab:spiritualist_corr} shows that MOTe has much higher absolute correlation than either F1 or Dice on all 14 pooled and individual tests. This is because, although Missing, Overlap or Trespass can have a perfect score but a range of CER values, the overall MOTe cannot; this is not the case for the F1 or Dice, which struggle to provide a sensible representation of quality at any value of CER.

Table \ref{tab:bounds_results_real} shows that both the full and simplified upper bound have high adherence, while the lower bound, $\max(\mathcal{M}, \mathcal{O}) \leq\mathrm{CER}$,  has notably lower values for both generated and real images. However, strong clustering around the lower bound makes it an effective proxy for parsing-induced CER. On the generated text, the lower bound has an $ R^2$ of 0.84 including PPDoc-M/S and $R^2=0.99$ excluding them. On The Spiritualist, the change is substantial, going from $-0.59$ to $0.83$ when PPDoc-M/S are removed, as they cause massive skew by predicting Trespassing bounding boxes that cover the entire page. We also see the distance distribution from the bounds can be skewed, with large differences between the mean and median, but this is not consistent.

When we ablate the experiment using simulated predictions, we find that F1 and Dice struggle to provide useful scores due to F1's granularity sensitivity and Dice's inability to separate instances. This results in F1 producing consistently low overall scores and correlations with CER, while Dice produces consistently high scores and correlations that switch sign depending on whether the predicted and GT columns match.  We see that MOTe's ability to act as a proxy for CER is sensitive to the amount of vertical vs horizontal Trespass, and the text recoverability assumption being violated. High relative vertical Trespass reduces the absolute correlation between MOTe and CER, as covering text before or after in a column has no impact on reading order despite breaking semantic boundaries. We found that the CER upper bound was breached predominantly when the text recoverability assumption is violated and text boxes are split down the middle. This difference between horizontal and vertical Trespass caused $\mathrm{CER}\geq \mathrm{max}(\mathcal{M}, \mathcal{O})$ to collapse in off-diagonal scenarios, while double-counting trespass such that $\mathrm{MOTe}=\mathcal{M}+ \mathcal{O}+2\mathcal{T}$ was more stable but performed at or below standard MOTe. Detail on the ablations can be found in Supplementary Material Section 4.

Given the findings on model correlation, we compared the ability of the metrics to pick the model with the least parsing-induced error for all combinations of model and image. Table \ref{tab:spiritualist_head_to_head} shows that MOTe performs very well, picking the right model 86\% of the time on the generated data and 98\% of the time on The Spiritualist. MOTe beats F1 and Dice by a substantial margin, outperforming F1 by 18 percentage points on the generated set and 19 percentage points on The Spiritualist. When $\max(\mathcal{M},\mathcal{O})$ is used, it outperforms MOTe only when cross-column Trespass is not a major factor; when it is, the score can drop very low. Confidence intervals are shown based on 10,000 bootstrap re-samples.

\begin{table*}
\caption{Spearman correlation of CER with each score for the five layout models on the synthetic pages (450 per model) and on The Spiritualist (49 per model). MOTe and $\max(\mathcal{M}, \mathcal{O})$ are errors, so positive values indicate agreement with CER; F1@0.5 and Dice are scores, so negative values indicate agreement with CER. PPDoc-M and -S return about two page-sized boxes per page on The Spiritualist, so the pooled correlation is also given without them.}
\label{tab:spiritualist_corr}
\centering
\begin{tabular}{llrrrr}
\toprule
 &  & MOTe & $\max(\mathcal{M}, \mathcal{O})$ & F1@0.5 & Dice \\
Dataset & Model &  &  &  &  \\
\midrule
\multirow[c]{7}{*}{Synthetic} & Pooled & 0.82 & 0.89 & -0.54 & -0.62 \\
 & Pooled excl. PPDoc-M/S & 0.83 & 0.94 & -0.51 & -0.49 \\
 & DocLayout-YOLO & 0.91 & 0.98 & -0.55 & -0.48 \\
 & DoclingLayoutHeron & 0.81 & 0.88 & -0.46 & -0.33 \\
 & PPDocLayout-L & 0.82 & 0.96 & -0.49 & -0.58 \\
 & PPDocLayout-M & 0.81 & 0.80 & -0.59 & -0.73 \\
 & PPDocLayout-S & 0.75 & 0.82 & -0.43 & -0.59 \\
\midrule
\multirow[c]{7}{*}{Spiritualist} & Pooled & 0.96 & -0.29 & -0.69 & -0.42 \\
 & Pooled excl. PPDoc-M/S & 0.97 & 0.89 & 0.01 & -0.57 \\
 & DocLayout-YOLO & 0.97 & 0.88 & 0.17 & -0.19 \\
 & DoclingLayoutHeron & 0.96 & 0.96 & 0.26 & -0.48 \\
 & PPDocLayout-L & 0.97 & 0.85 & -0.23 & -0.84 \\
 & PPDocLayout-M & 0.66 & 0.16 & -0.38 & -0.08 \\
 & PPDocLayout-S & 0.47 & -0.24 & 0.08 & 0.08 \\
\bottomrule
\end{tabular}
\end{table*}

\begin{table*}
    \centering
    \small
    \begin{tabular}{llcccccc}
    \toprule
        & & \multicolumn{3}{c}{\textbf{All models}} & \multicolumn{3}{c}{\textbf{excl.\ PPDoc-M/S}} \\
        \cmidrule(lr){3-5} \cmidrule(lr){6-8}
        \textbf{Data} & \textbf{Bound} & Adh. & Mean & Median & Adh. & Mean & Median \\
        \midrule
        Synthetic    & $\mathrm{CER}\leq \mathcal{M}+\mathcal{O}+ 2\mathcal{T}$ & 0.999 & 0.340 & 0.089 & 0.999 & 0.350 & 0.129 \\
                     & $\mathrm{CER}\leq \mathrm{MOTe}$                          & 0.991 & 0.168 & 0.055 & 0.998 & 0.184 & 0.081 \\
                     & $\mathrm{CER}\geq \max(\mathcal{M}, \mathcal{O})$          & 0.817 & 0.019 & $-0.003$ & 0.833 & $-0.001$ & $-0.003$ \\
        \midrule
        Spiritualist & $\mathrm{CER}\leq \mathcal{M}+\mathcal{O}+ 2\mathcal{T}$ & 1.000 & 0.314 & 0.080 & 1.000 & 0.066 & 0.053 \\
                     & $\mathrm{CER}\leq \mathrm{MOTe}$                          & 0.620 & 0.018 & 0.035 & 0.986 & 0.055 & 0.050 \\
                     & $\mathrm{CER}\geq \max(\mathcal{M}, \mathcal{O})$          & 0.641 & 0.294 & 0.016 & 0.401 & $-0.016$ & $-0.030$ \\
        \bottomrule
    \end{tabular}
    \caption{Adherence (fraction of pages satisfying each bound) and the mean and median signed
    distance from the bound, pooled over the five layout models and without PPDoc-M and -S.}
    \label{tab:bounds_results_real}
\end{table*}

\begin{table*}
\centering
\caption{Head-to-head agreement with CER on the synthetic pages and on The Spiritualist, with and without PPDoc-M and -S: over every page and pair of models whose CER differs by more than 0.01, the fraction where the score prefers the model, with 95\% CIs from a page-level bootstrap (10,000 replicates).}
\label{tab:spiritualist_head_to_head}
\begin{tabular}{lcccc}
\toprule
 & \multicolumn{2}{c}{\textbf{Synthetic}} & \multicolumn{2}{c}{\textbf{Spiritualist}} \\
 & \textbf{All} & \textbf{excl.\ M/S} & \textbf{All} & \textbf{excl.\ M/S} \\
\midrule
MOTe & 0.856 [0.84, 0.87] & 0.859 [0.84, 0.88] & \textbf{0.986} [0.97, 1.00] & \textbf{0.977} [0.95, 1.00] \\
$\max(\mathcal{M}, \mathcal{O})$ & \textbf{0.938} [0.93, 0.95] & \textbf{0.965} [0.95, 0.98] & 0.312 [0.26, 0.37] & 0.744 [0.66, 0.82] \\
F1@0.5 & 0.679 [0.66, 0.70] & 0.654 [0.62, 0.69] & 0.792 [0.75, 0.83] & 0.541 [0.46, 0.62] \\
Dice & 0.768 [0.75, 0.78] & 0.680 [0.65, 0.71] & 0.347 [0.28, 0.41] & 0.617 [0.52, 0.71] \\
\midrule
Decisions & 4041 & 1111 & 430 & 133 \\
\bottomrule
\end{tabular}
\end{table*}

\subsection{Model and Dataset Comparison}
\label{sect:model_data_comp}

Having demonstrated the sensitivity of the F1 to granularity differences, and the strength of the MOTe in representing parsing-induced character errors, we now see how real models and data can be interpreted through this lens. Tables \ref{tab:res_spiritualist}, \ref{tab:res_ncse}, \ref{tab:res_HNLA2013}, and \ref{tab:res_doclaynet} show the results of comparing the 5 models across four datasets aggregated to dataset level using the mean. The tables show that while the scores can be similar on average, the traditional metrics provide no insight into why the score is high or low. In some cases, results can diverge dramatically, for example The Spiritualist has ground truth at line level, but the models predict closer to paragraph level, so IoU based metrics suggest very poor model performance, whilst MOTe suggests they did very well. Heron performed much better than the other models on DocLayNet as this was a core part of it's training data. 

A substantial difference when using the MOTe is that the CER that is a direct result of the parsing models can be estimated using the derived bounds. This is shown in the last column of the tables as P-CER (Parsing CER), and represents the irreducible character error before OCR has taken place. However, it should be noted that Section \ref{sec:res_cor_cer}, showed that whilst the upper bound is quite firm, the lower bound is more like a regression acting as a proxy for CER.

\begin{figure*}
    \centering
    \includegraphics[width=0.8\linewidth]{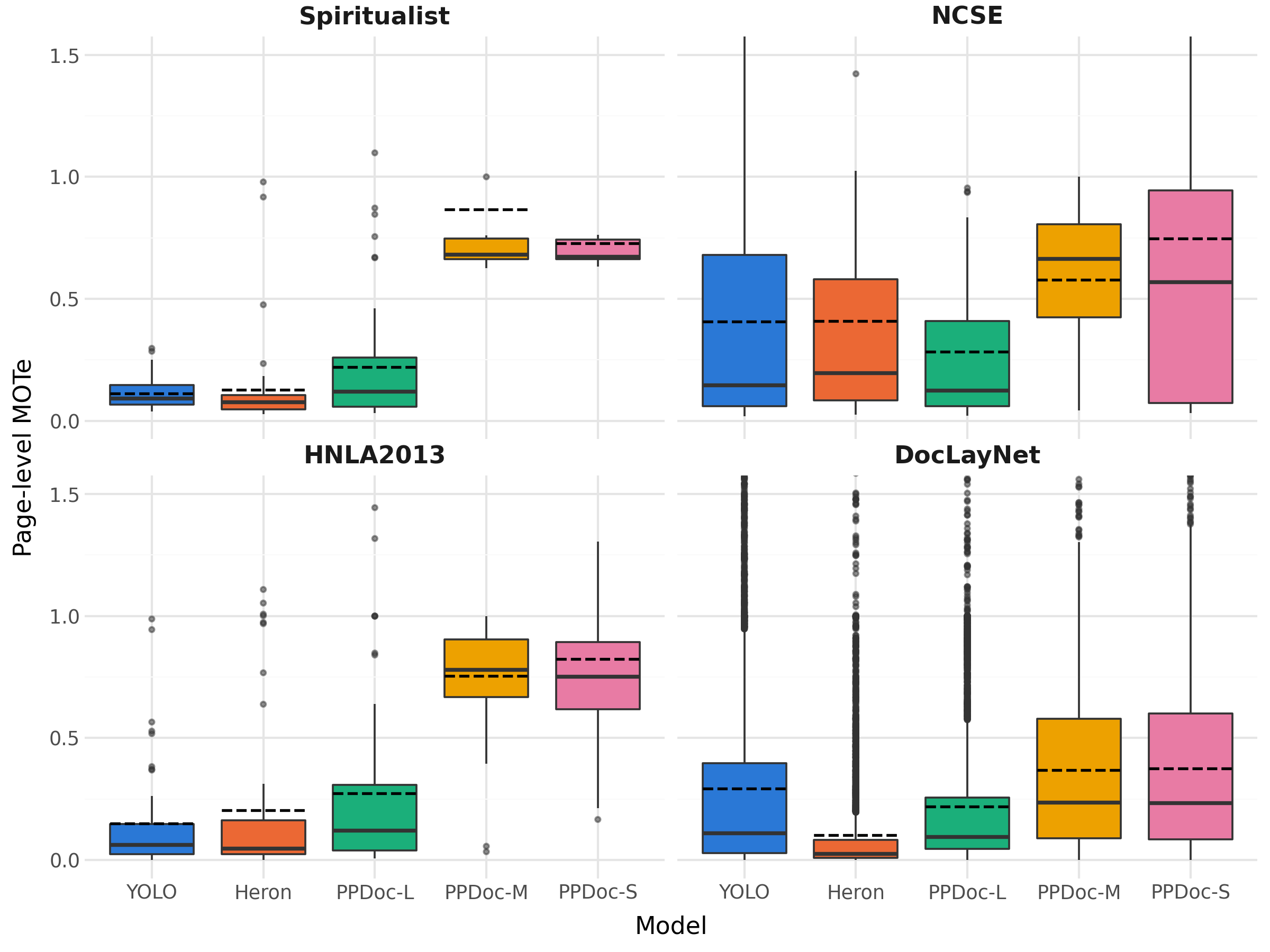}
    \caption{Similar to CER, MOTe errors are unbounded, so the mean can be skewed, it can therefore be valuable to look at the median and inter-quartile range to understand model performance differences.}
    \label{fig:mote_boxplot_by_dataset}
\end{figure*}

As MOTe only has a lower bound it can, similar to CER, produce skewed distributions, this can be valuable for interpreting model differences. As the tables only show the mean results we plot the results for all datasets and models in a box plot shown in Figure \ref{fig:mote_boxplot_by_dataset}. The Figure shows that there can be substantial differences between the mean and median, with Yolo and Heron swapping position for first place on The Spiritualist and HNLA2013 datasets.

As DocLayNet had such good overall performance despite not having SSU, we re-ran the NCSE and HNLA2013 datasets without SSU labelling. As discussed in Section \ref{sec:motivation_ssu}, this means that instead of using the semantic structure, the MOTe has to fall back on the same physical structure used by the F1 and mAP. We found that removing the SSU added only 0.05 Trespass to the NCSE on average and 0.06 to HNLA2013. Removing the poorly performing PPDoc-M and S meant that Trespass increased by only about 0.03 on average. 

\begin{figure*}
    \centering
    \includegraphics[width=\linewidth]{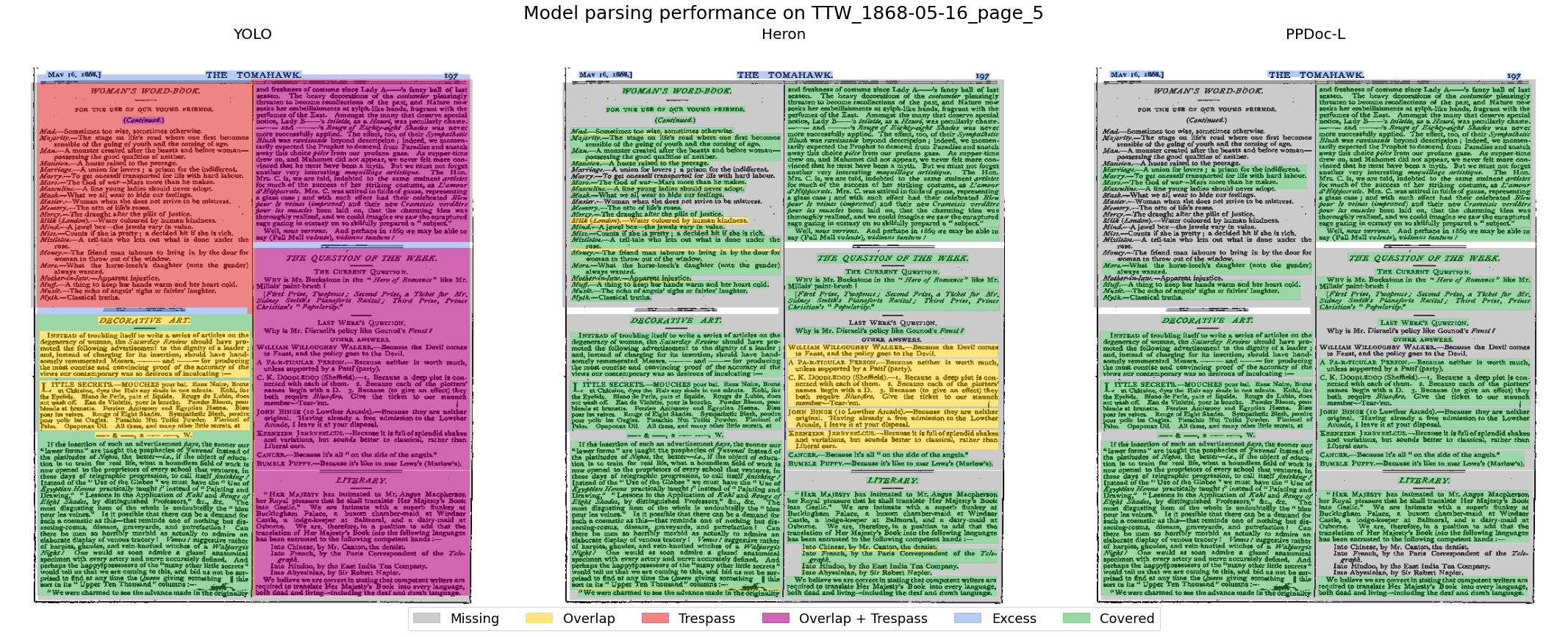}
    \caption{It is clear in visual analysis that the models can behave very differently on the same image, and have very different failure modes.}
    \label{fig:example_page_with_labels}
\end{figure*}

The value of the decomposition is shown at page level in Figure \ref{fig:example_page_with_labels} and Table \ref{tab:model-metrics}. It is clear that the models have very different behaviour and that MOTe demonstrates higher pragmatic competence tracking visible performance and failures (e.g. extreme Trespass and Overlap on the example page) more faithfully than IoU/F1. It is for this reason the MOTe is able to select the model which generates the least parsing-induced character errors as shown in Section \ref{sec:res_cor_cer}. The relatively large amount of white space labelled as Missing is a visual example of granularity sensitivity explored in Section \ref{sec:gran_robust}, and explains why the MOTe scores are higher than one would expect for Heron and PPDoc-L when looking at the image. It also shows the value of the many-to-one assignment when MOTe is compared to the F1 scores which are close to 0.

\begin{table*}[htbp]
\centering
\caption{Model results for the Spiritualist. P-CER is the estimated parsing-induced CER.}
\label{tab:res_spiritualist}
\begin{tabular}{lccccccccc}
\toprule
\textbf{Model} & \textbf{MOTe} & \textbf{Missing} & \textbf{Overlap} & \textbf{Trespass} & \textbf{Excess} & \textbf{IoU} & \textbf{mAP} & \textbf{F1@50} & \textbf{P-CER} \\
\midrule
Heron & 0.13 & 0.05 & 0.08 & \textbf{0.00} & 0.09 & \textbf{0.11} & 0.00 & \textbf{0.06} & 0.10--0.13 \\
PPDoc-L & 0.22 & 0.17 & 0.03 & 0.03 & \textbf{0.08} & 0.09 & 0.00 & 0.04 & 0.18--0.25 \\
PPDoc-M & 0.87 & 0.03 & 0.10 & 0.74 & 0.33 & 0.00 & 0.00 & 0.00 & 0.13--1.61 \\
PPDoc-S & 0.73 & \textbf{0.01} & \textbf{0.02} & 0.70 & 0.29 & 0.00 & \textbf{0.00} & 0.00 & 0.03--1.43 \\
YOLO & \textbf{0.11} & 0.05 & 0.06 & 0.00 & 0.10 & 0.10 & 0.00 & 0.04 & 0.07--0.12 \\
\bottomrule
\end{tabular}
\end{table*}

\begin{table*}[htbp]
\centering
\caption{Model results for NCSE. P-CER is the estimated parsing-induced CER.}
\label{tab:res_ncse}
\begin{tabular}{lccccccccc}
\toprule
\textbf{Model} & \textbf{MOTe} & \textbf{Missing} & \textbf{Overlap} & \textbf{Trespass} & \textbf{Excess} & \textbf{IoU} & \textbf{mAP} & \textbf{F1@50} & \textbf{P-CER} \\
\midrule
Heron & 0.41 & 0.13 & 0.28 & \textbf{0.00} & \textbf{0.03} & \textbf{0.75} & \textbf{0.56} & 0.51 & 0.37--0.41 \\
PPDoc-L & \textbf{0.28} & 0.20 & 0.05 & 0.03 & 0.03 & 0.70 & 0.38 & \textbf{0.57} & 0.25--0.31 \\
PPDoc-M & 0.58 & 0.24 & \textbf{0.05} & 0.29 & 0.12 & 0.36 & 0.09 & 0.26 & 0.26--0.87 \\
PPDoc-S & 0.75 & 0.13 & 0.21 & 0.41 & 0.12 & 0.44 & 0.14 & 0.29 & 0.33--1.16 \\
YOLO & 0.41 & \textbf{0.08} & 0.17 & 0.16 & 0.06 & 0.72 & 0.37 & 0.55 & 0.21--0.57 \\
\bottomrule
\end{tabular}
\end{table*}

\begin{table*}[htbp]
\centering
\caption{Model results for HNLA2013. P-CER is the estimated parsing-induced CER.}
\label{tab:res_HNLA2013}
\begin{tabular}{lccccccccc}
\toprule
\textbf{Model} & \textbf{MOTe} & \textbf{Missing} & \textbf{Overlap} & \textbf{Trespass} & \textbf{Excess} & \textbf{IoU} & \textbf{mAP} & \textbf{F1@50} & \textbf{P-CER} \\
\midrule
Heron & 0.20 & \textbf{0.02} & 0.18 & \textbf{0.01} & \textbf{0.13} & \textbf{0.81} & \textbf{0.62} & \textbf{0.82} & 0.18--0.21 \\
PPDoc-L & 0.27 & 0.17 & 0.05 & 0.05 & 0.16 & 0.61 & 0.46 & 0.74 & 0.22--0.32 \\
PPDoc-M & 0.75 & 0.31 & \textbf{0.00} & 0.44 & 0.45 & 0.06 & 0.01 & 0.02 & 0.31--1.20 \\
PPDoc-S & 0.82 & 0.19 & 0.08 & 0.55 & 0.48 & 0.09 & 0.02 & 0.06 & 0.27--1.38 \\
YOLO & \textbf{0.15} & 0.03 & 0.07 & 0.05 & 0.15 & 0.74 & 0.44 & 0.68 & 0.08--0.20 \\
\bottomrule
\end{tabular}
\end{table*}

\begin{table*}[htbp]
\centering
\caption{Model results for DocLayNet. P-CER is the estimated parsing-induced CER.}
\label{tab:res_doclaynet}
\begin{tabular}{lccccccccc}
\toprule
\textbf{Model} & \textbf{MOTe} & \textbf{Missing} & \textbf{Overlap} & \textbf{Trespass} & \textbf{Excess} & \textbf{IoU} & \textbf{mAP} & \textbf{F1@50} & \textbf{P-CER} \\
\midrule
Heron & \textbf{0.10} & 0.05 & \textbf{0.04} & \textbf{0.02} & \textbf{0.03} & \textbf{0.85} & \textbf{0.76} & \textbf{0.90} & 0.08--0.12 \\
PPDoc-L & 0.22 & 0.09 & 0.05 & 0.08 & 0.07 & 0.66 & 0.42 & 0.71 & 0.13--0.30 \\
PPDoc-M & 0.37 & 0.22 & 0.06 & 0.09 & 0.07 & 0.51 & 0.29 & 0.61 & 0.26--0.46 \\
PPDoc-S & 0.38 & 0.23 & 0.05 & 0.09 & 0.07 & 0.47 & 0.29 & 0.61 & 0.27--0.47 \\
YOLO & 0.29 & \textbf{0.04} & 0.15 & 0.11 & 0.08 & 0.72 & 0.49 & 0.76 & 0.17--0.40 \\
\bottomrule
\end{tabular}
\end{table*}

\begin{table*}[htbp]
\centering
\caption{Per-model metrics for \texttt{TTW\_1868-05-16\_page\_5}, bold is best in column}
\label{tab:model-metrics}
\begin{tabular}{lccccccc}
\toprule
\textbf{Model} & \textbf{MOTe} & \textbf{Missing} & \textbf{Overlap} & \textbf{Trespass} & \textbf{Excess} & \textbf{mIoU} & \textbf{F1} \\
\midrule
YOLO & 1.60 & \textbf{0.02} & 0.59 & 0.98 & 0.10 & 0.25 & 0.00 \\
Heron & \textbf{0.34} & 0.23 & 0.12 & \textbf{0.00} & \textbf{0.02} & 0.44 & 0.06 \\
PPDoc-L & 0.35 & 0.35 & \textbf{0.00} & \textbf{0.00} & 0.02 & \textbf{0.45} & \textbf{0.12} \\
\bottomrule
\end{tabular}
\end{table*}

Table \ref{tab:pred_gt_ratio_spearman} explores the impact of number of predictions, we compare the absolute log ratio ($|\log_2 \text{ratio}|$) of prediction to ground truth. This value is symmetric and so counts predictions above or below the ground truth number equally. We find that MOTe, IoU, mAP, and F1 all worsen with divergence from the number of ground truth regions (MOTe is positively correlated and IoU, mAP, and F1 are negatively correlated), with MOTe the weakest or joint weakest in absolute correlation on all three datasets. This is because MOTe is not directly dependent on matching the number of predictions unlike F1, IoU or mAP. We see that the MOTe correlates more strongly than its components, this is because it penalises the bad parsing that can co-occur with large divergences from the ground truth region counts. In contrast, the F1 penalises all count divergence and cannot disambiguate parsing has error from granularity mismatch (see Section \ref{sec:meas_gran_robust}).

\begin{table*}[htbp]
\centering
\caption{Page-level Spearman correlation between $|\log_2 \text{Ratio}|$ of predicted to ground truth regions and each metric, per dataset (pages pooled over all five models). MOTe and its components are errors, so positive values indicate worse performance with divergence; IoU, mAP and F1@50 are scores, so negative values indicate worse performance.}
\label{tab:pred_gt_ratio_spearman}
\begin{tabular}{lccccccc}
\toprule
\textbf{Dataset} & \textbf{MOTe} & \textbf{Missing} & \textbf{Overlap} & \textbf{Trespass} & \textbf{IoU} & \textbf{mAP} & \textbf{F1@50} \\
\midrule
DocLayNet & 0.59 & 0.29 & 0.13 & 0.31 & -0.67 & -0.67 & -0.83 \\
HNLA2013 & 0.83 & -0.21 & -0.46 & 0.59 & -0.88 & -0.88 & -0.93 \\
NCSE & 0.51 & -0.18 & 0.11 & 0.24 & -0.51 & -0.53 & -0.82 \\
\bottomrule
\end{tabular}
\end{table*}

\section{Discussion}
\label{sec:discussion}
We find that traditional DLA metrics like the F1 are not appropriate for evaluating page parsing quality. When tested under different granularity mismatches, in Section \ref{sec:gran_robust}, the F1 diverged from the true score by 98 percentage points compared to the MOTe diverging by only 9 percentage points. In Section \ref{sect:f1_cer}, we also showed that the F1 could produce a wide range of CER values even when F1=1, whilst in contrast when MOTe=0, CER = 0. We demonstrated this further in Section \ref{sec:res_cor_cer}, showing that MOTe was much more strongly correlated with parsing-induced character errors ($\left| \rho \right|= 0.82$) than either F1 ($\left| \rho \right|= 0.54$) or Dice ($\left| \rho \right|= 0.62$) on synthetic data. This relationship with CER translated into decision making with the MOTe choosing the model which would produce the lowest parsing-induced character errors 86\% of the time, compared to F1 at 68\% and Dice at 77\%, on synthetic data. Repeating these tests on the 49-page 19th-century newspaper The Spiritualist, MOTe outperformed the other metrics by a similar magnitude. This difference reflects MOTe’s higher semantic competence, in that it is able to disentangle and accurately penalise different error types, given the relationship between predictions and the ground truth, as well as the relationship between the predictions themselves. This gap in performance between the MOTe and the other metrics, shows that the MOTe has higher pragmatic competence, and that unlike the F1 and Dice, the integrity of the model-data-metric triad is maintained across a wider range of scenarios. This is because MOTe was developed for the natively 2D tessellation structures that commonly occur in human-readable documents.

The consequences of these differences were shown in Section \ref{sect:model_data_comp}, where MOTe error frequently disagreed with the other metrics both in terms of best performing model, and the overall ranking of the models. However care has to be taken with the meaning of `best', as MOTe is similar to the CER in that it can be skewed by a few outsized errors, so practitioners need to consider aggregation method depending on their use and goals. In addition, we showed MOTe error provides interpretability into model errors at both dataset and individual image level, which allow practitioners to make decisions based on their own use case. This has broader implications for the field of DLA, as benchmarks which use the F1 as a scoring tool may be ranking models on a basis which is unrelated to their downstream use. The lack of decomposability also lumps different types of performance together preventing practitioners choosing the model most appropriate for their task which might be different from raw ranking. Both cases result in sub-optimal model selection which affects the wider DU community.

On a practical note, datasets like The Spiritualist, which has line-level bounding boxes, require SSU labelling for MOTe to evaluate most models fairly. However, we found that NCSEv2 and HNLA2013 derive most of their benefit from MOTe without explicit SSU labelling. As such, we recommend that practitioners first try the MOTe without SSU; if there is substantial Trespass, visualise the results (use \verb|visualize_mote_states| from the \verb|mote_error| library), and, if predictions cover multiple semantically linked ground truth regions, perform a trial SSU labelling and re-evaluate.

Future work could integrate the mathematical clarity of the MOTe with the detailed profile-driven approaches of \cite{clausner_scenario_2011} and \cite{heo_led_2026}, could provide an approach to evaluation with a clear conceptual thread from pixel level to discrete error types without the issues related to IoU based metrics. This would allow error profiles highly specific to the practitioner’s use case while remaining comparable across practitioners and models by using MOTe as a lower level of abstraction. Identifying where and how PAGE, ALTO, hOCR, TEI, and custom dataset schemas diverge, and at what level of abstraction they can be reconciled using a taxonomy of labelling schemas, would pair well with this integration. This is because practitioners using the Clausner/LED-style profile layer would need such a schema-mapping to provide a principled basis for comparability. Relatedly, explicitly extracting the hierarchical semantic structure from printed materials and creating a hierarchical SSU could further increase flexibility in the model-metric relationship.

Although we created the class system, we did not apply it; a follow-up analysis exploring class confusion and the impact of fine-tuning on semantic coherence would expand the framework’s understanding and scope. Similarly, we found several cases where the difference between horizontal and vertical Trespass affected the relationship with CER; breaking these out could provide more flexibility and value depending on practitioners’ end use. Finally, Supplementary Material Section 5 provides an example of MOTe applied to natural scenes and its relation to Panoptic Quality; a more thorough analysis of MOTe’s general applications in machine vision would be valuable.

\section{Conclusions}
This paper introduces two new concepts: MOTe error and the SSU labelling mechanism. Together, they create a decomposable error framework that is robust to text-label granularity and has high semantic competence in that it is able to express the relationship between the predictions and the ground truth, as well as the relationship of predictions with each other. Together, this creates a metric with high pragmatic competence that is broadly agnostic to labelling-schema and model training-data granularity, ensuring a robust model-data-metric relationship across a broad range of scenarios.

This robustness was demonstrated by MOTe dropping only 9 percentage points on average during granularity mismatch testing, compared with a 98-point drop for F1. MOTe also correctly chose the model that produced the fewest parsing-induced character errors in an image 86\% of the time, compared with Dice (77\%) and F1 (68\%).

Due to the substantial performance differences in tests directly applicable to common DU tasks, we suggest that the F1, Dice, and related metrics are not appropriate tools for measuring the quality of page parsing models and that MOTe error should be used preferentially. To support this, we release the \verb|mote_error| library, which includes code for labelling, analysing, and visualising data within the MOTe framework.

\section{Acknowledgements}

We would like to thank Dr Christian Clausner of the University of Salford for his support in relation to the HNLA2013 dataset. In addition, Wenbo Zhang for her valuable discussion on Wittgenstein and conceptual relativism.

\section{Data Availability}

The \verb|mote_error| library is available from \url{https://github.com/THE-3TC/mote_error} or can be installed from pip using \verb|pip install mote_error|. The NCSE dataset is available from \url{https://doi.org/10.5522/04/28381610}. The access to HNLA2013 was granted by \url{https://www.primaresearch.org}. The DocLayNet dataset can be downloaded from Huggingface Hub.

\bibliographystyle{unsrtnat}
\bibliography{citations}

\setcounter{section}{0}
\renewcommand{\thesection}{S\arabic{section}}
\renewcommand{\thesubsection}{\thesection.\arabic{subsection}}
\renewcommand{\theequation}{S\arabic{equation}}
\setcounter{equation}{0}
\renewcommand{\thefigure}{S\arabic{figure}}
\setcounter{figure}{0}
\renewcommand{\thetable}{S\arabic{table}}
\setcounter{table}{0}

\section{Deriving the upper bound of the MOTe}

The below sections provide the derivation for the upper bound of the MOTe

\subsection{Overlap}
Overlap is minimised at 0 whilst the maximum occurs when all $n$ predictions cover the entire ground truth region shown in $M^\mathrm{S}$, this results in

\begin{equation}
\mathcal{O}_\mathrm{max} = n \frac{A^\mathrm{S}}{A^\mathrm{S}} - 1= n - 1
\end{equation}

\subsection{Trespass}

Similar to Overlap, Trespass is minimised at 0, however the upper limit is different. The maximum Trespass of an individual prediction occurs when the prediction covers the entire page and is shown as

\begin{equation}
    t_{\mathrm{max}} = \frac{A^\mathrm{S} - A^\mathrm{S}_\mathrm{max}}{A^\mathrm{S}} = 1 - \frac{A^\mathrm{S}_\mathrm{max}}{A^\mathrm{S}}
\end{equation}

Where $A^\mathrm{S}_\mathrm{max}$ is the area of the largest SSU such that for a page of $m$ SSU $ m\frac{A^\mathrm{S}_\mathrm{max}}{A^\mathrm{S}}\geq 1$. As such, as $m \rightarrow \infty$, and the area of the largest SSU goes to zero we see that the Trespass of an individual prediction approaches 1

\begin{equation}
    \underset{A^\mathrm{S}_\mathrm{max}\to 0}{\mathrm{lim}} \; 1 - \frac{A^\mathrm{S}_\mathrm{max}}{A^\mathrm{S}} = 1 - \frac{0}{A^\mathrm{S}}=1
\end{equation}

This results in a theoretical upper bound of

\begin{equation}
    \mathcal{T}_\mathrm{max\:theoretical} = \sum_{j=1}^{n} 1 = n
\end{equation}

for $n$ predictions on $m$ SSU. As all predictions will be assigned to the largest or joint largest SSU.

However, under non-zero SSU sizes the maximum possible Trespass is 

\begin{equation}
    \mathcal{T}_\mathrm{max} = n(1 - \frac{A^\mathrm{S}_\mathrm{max}}{A^\mathrm{S}})
\end{equation}

\subsection{Upper bound of the MOTe}

Given that the previously derived upper bounds for Overlap and Trespass require covering the entire page, it means they occur when Missing is minimised. As such, we can find the upper bound of the MOTe using the MOTe formula

\begin{equation}
\mathrm{MOTe} = \mathcal{M} + \mathcal{O} + \mathcal{T}
\end{equation}

Substituting in the bounds
\begin{equation}
   \mathrm{MOTe}_{\mathrm{limit}} = 0 + (n-1) + (n)
\end{equation}

and simplifying the theoretical upper limit is
\begin{equation}
  \mathrm{MOTe}_{\mathrm{max\:theoretical}} = 2n-1
\end{equation}

Whilst the practical upper limit for a page of finite sized SSU is

\begin{equation}
  \mathrm{MOTe}_{\mathrm{max\:practical}} = 2n - 1 - n\frac{A^\mathrm{S}_\mathrm{max}}{A^\mathrm{S}}
\end{equation}

\section{Additional diagnostics}

Beyond the MOTe $F_{1}$ score described in Section 3.7 of the manuscript, we can calculate the total fraction of each metric made up of class $k$. For convenience we first define the class-$k$ prediction fraction
\begin{equation}
    \widehat{M}^{\mathrm{p}}_k \equiv M^{\mathrm{p}}_k / M^{\mathrm{p}},
\end{equation}
which is the pixelwise share of predicted coverage attributed to class $k$. Using this, the class-wise Coverage is

\begin{equation}
    \mathcal{C}^{\mathrm{share}}_k = \frac{\sum_{x,y} M^{\mathrm{S}} \odot \widehat{M}^{\mathrm{p}}_k}{\mathcal{C}\cdot A^{\mathrm{S}}}
\end{equation}

Which is the sum of the area of the true coverage (positive predictions) of class $k$ as a fraction of all coverage. Class wise Overlap is

\begin{equation}
    \mathcal{O}^{\mathrm{share}}_k = \frac{\sum_{x,y} (M^{\mathrm{S}} \odot \widehat{M}^{\mathrm{p}}_k)\odot (M^{\mathrm{p}} - M^{\mathrm{p,b}}) }{\mathcal{O}\cdot A^{\mathrm{S}}}
\end{equation}

The sum of the product of Coverage matrix of class $k$ and the matrix of overlapping predictions, as a fraction of all Overlap. Finally class based Trespass is

\begin{equation}
    \mathcal{T}^{\mathrm{share}}_k = \frac{\sum_{j \in k} \sum_{x,y} M^{\mathrm{S}}_{\backslash i(j)} \odot M^{\mathrm{p}}_j}{\mathcal{T} \cdot A^{\mathrm{S}}}
\end{equation}
Where we simply sum the total area of Trespass of each prediction of class $k$ across all predictions of class $k$, and divide by the total Trespass. Trespass does not need pixel share normalisation as it is relative to the prediction which can only be a single class and the ground truth so has a direct mapping which handles multiple Trespasses naturally.

In addition to the fractional make up of the MOTe components, the relational nature of the MOTe error allows us to create an asymmetric confusion matrix between all $K$ classes for each of the Coverage, Overlap, and Trespass. This is normalised by the total area of prediction of class $k$ such that $A^{\mathrm{P}}_k = \sum_{x,y}M_k^{p,b}$ this means the confusion matrix is from the prediction perspective which supports model diagnostics as opposed to more traditional confusion matrices. 

\begin{equation}
    \mathcal{C}_{k,l} = \frac{\sum_{x,y} M^{\mathrm{S}}_l \odot  M^\mathrm{p,b}_k}{A^{\mathrm{P}}_k}
\end{equation}

Diagonal entries $\mathcal{C}_{k,k}$ give correct within-class coverage. Off-diagonal entries indicate cross-class coverage indicative of classification errors.

We define the binary Overlap mask as the pixels covered by more than one prediction,
\begin{equation}
    M^{\mathcal{O}} = 1\left[M^{\mathrm{p}} > 1\right].
\end{equation}

The Overlap confusion matrix is then
\begin{equation}
    \mathcal{O}_{k,l} = \frac{\sum_{j \in k} \sum_{x,y} M^{\mathcal{O}} \odot M^{\mathrm{p}}_j \odot M^{\mathrm{S}}_l}{A^{\mathrm{O}}_k},
\end{equation}

Which is the intersection of the pixels with overlap, and prediction of class $k$ and ground truth of class $l$ summed across all predictions of class $k$ and normalised by

\begin{equation}
    A^{\mathrm{O}}_k = \sum_{j \in k} \sum_{x,y} M^{\mathcal{O}} \odot M^{\mathrm{p}}_j \odot M^{\mathrm{S}},
\end{equation}

which is the sum of the intersection of all overlapping pixels and all predictions of class $k$ and all ground truth regions. Diagonal entries arise where predictions overlap within ground truth of their own class. Off-diagonal entries are either classification errors or Trespass errors that also overlap.

A Trespass confusion matrix can also be created that shows the fraction of total Trespass of predictions of class $k$ onto ground truth of class $l$

\begin{equation}
    \mathcal{T}_{k,l} = \frac{\sum_{j \in k} \sum_{x,y} M^{\mathrm{S}}_{l \backslash i(j)} \odot M^{\mathrm{p}}_j}{A^{\mathrm{P}}_k}
\end{equation}

Diagonal entries show where a prediction has Trespassed against its own class, however by definition a prediction cannot Trespass against its own assigned SSU.

\section{Example Case studies}

This section provides a series of simple examples based on the limerick figure from the main text.

\subsection{Method}
To provide intuitive examples of the introduced metrics in practice we will use a set of simple case studies based on the limericks used in the paper. 

We will first demonstrate the difference in apparent performance when using SSU based analysis in comparison to IoU. To demonstrate the granularity robustness of the SSU relative to traditional metrics, we consider the case if the limericks were labelled with ground truth at line level (resulting in 18 regions), but the predictions, whilst perfectly encompassing the text, were at paragraph level (7 regions) or vice versa. From an end-to-end perspective there is no impact on the extracted text, and as such both metrics should provide perfect scores.

We will then use the limerick example presented with different types of parsing issues and quantify the impact in terms of MOTe, F1, IoU, and CER.

Finally we will provide a simple calculation of class based analysis for the example

\subsection{Results}

\subsubsection{SSU and IoU}
\label{sec:ssu_iou}

We quantify the impact on measured performance of perfect page parsing, but with substantial granularity differences between model and ground truth. As shown in Table \ref{tab:example_granularity}, the traditional metrics produce highly misleading results, suggesting the model has performed very poorly. In both scenarios, the F1 is 0.32, and its sub-metrics mirror each other across the cases. 

In contrast, the MOTe error is largely robust to differences in granularity. When the ground truth is at line level, it has a perfect error of 0; when the ground truth is at paragraph level, it has an error of 0.16. This drop in apparent performance is because the short, irregular lines of poetry result in voids within the ground truth paragraph bounding box (paragraph boxes shown in \autoref{fig:additional_parsing_issues} c), producing artificially higher Missing for the void regions. This shows that when under loose-labelling, which creates significant white-space voids, the MOTe loses accuracy, but relative to the F1, the MOTe is much more reliable.

These granularity examples are shown in Supplementary Material Section 3.

Overall this case study provides a concrete example of how labelling differences between prediction and ground truth can degrade the triad relationship unless metrics have high pragmatic competence.

\begin{table}
    \centering
\begin{tabular}{lcc}
\toprule
Metric & \thead{GT: Line \\ Pred: Para} & \thead{GT: Para \\ Pred: Line}  \\
\midrule
True Positive & 4 & 4 \\
False Positive & 3 & 14 \\
False Negative & 14 & 3 \\
Precision & 0.57 & 0.22 \\
Recall & 0.22 & 0.57 \\
F1 & 0.32 & 0.32 \\
Mean IoU & 0.35 & 0.60 \\
MOTe error & 0.00 & 0.16 \\
\bottomrule
\end{tabular}
    \caption{Comparison of MOTe vs traditional metrics on granularity-mismatched predictions from the limerick example. Traditional metrics suggest almost complete failure despite perfect parsing. However, the MOTe correctly identifies excellent performance.}
    \label{tab:example_granularity}
\end{table}

\subsubsection{The MOTe}

\begin{figure}[htbp]
  \centering
  \begin{subfigure}[b]{0.49\textwidth}
    \includegraphics[width=\linewidth]{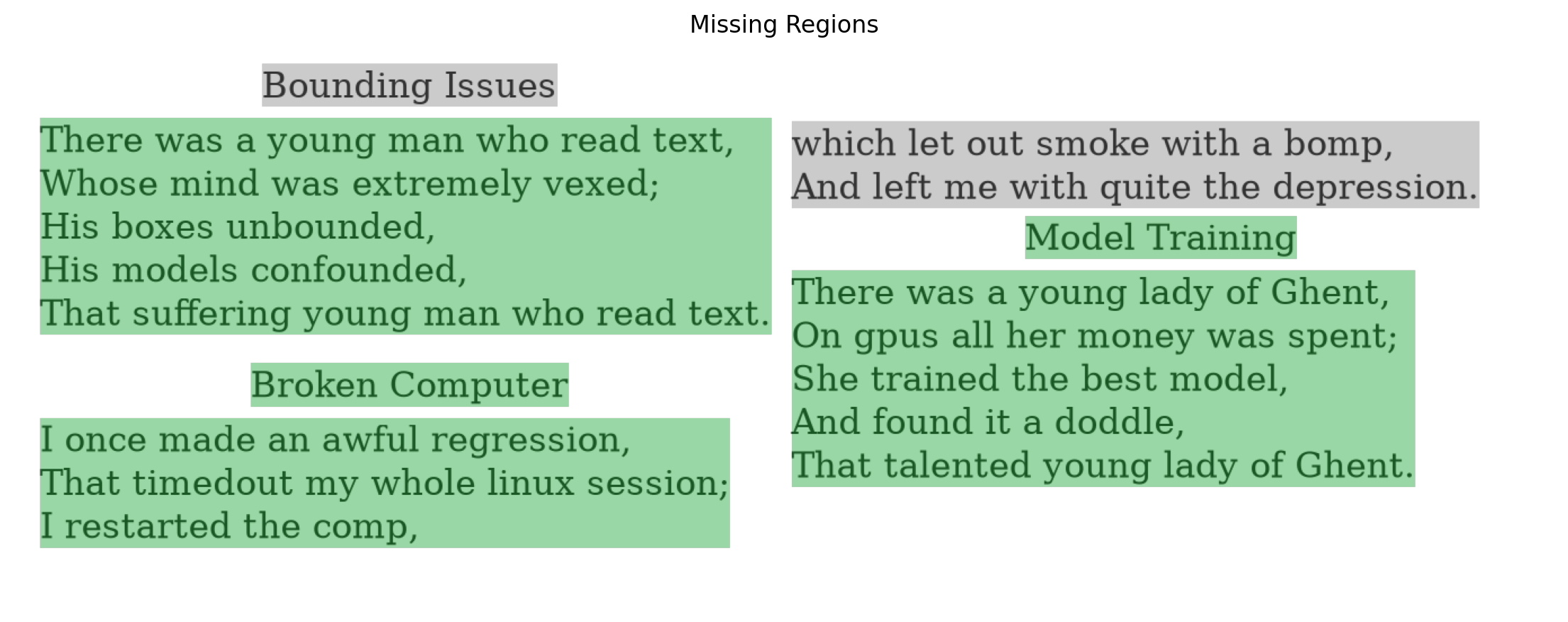}
    \caption{Missing some ground truth}
    \label{fig:a}
  \end{subfigure}
  \hfill
  \begin{subfigure}[b]{0.49\textwidth}
    \includegraphics[width=\linewidth]{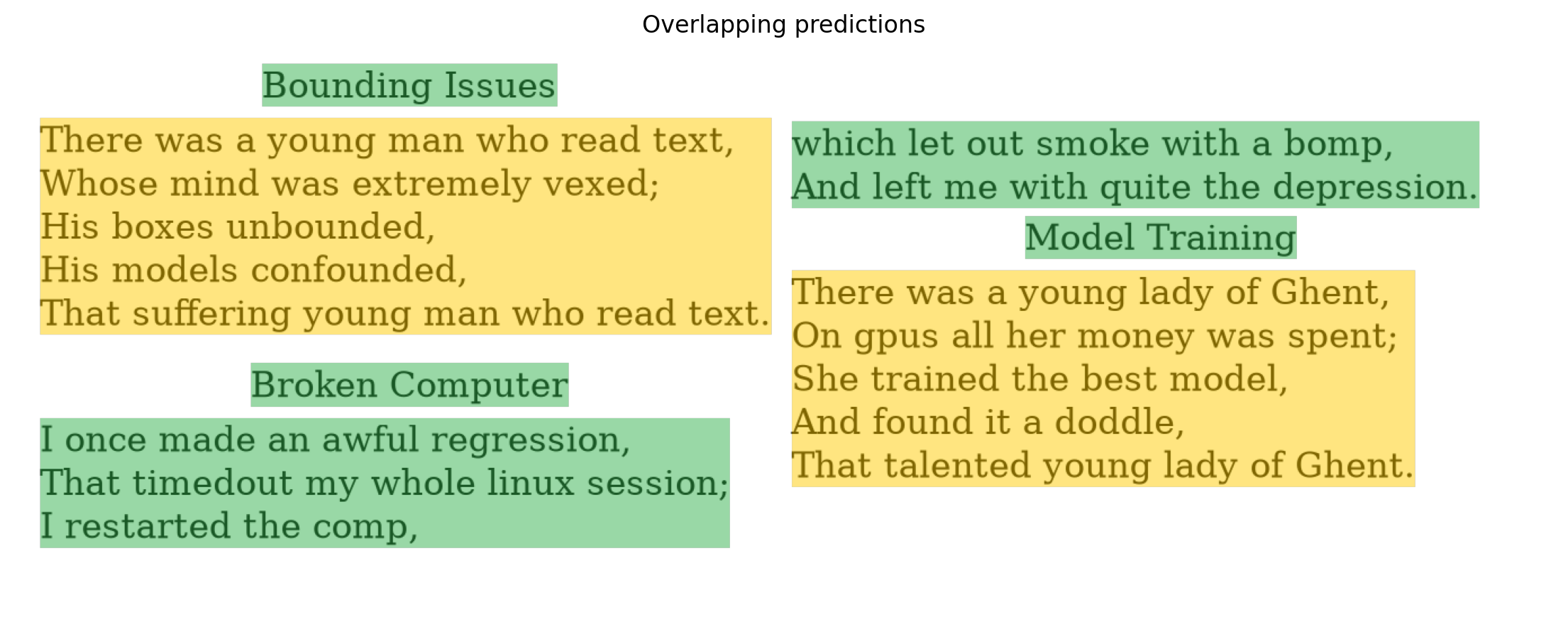}
    \caption{Overlapping on ground truth}
    \label{fig:b}
  \end{subfigure}

  \vspace{0.5em}

  \begin{subfigure}[b]{0.49\textwidth}
    \includegraphics[width=\linewidth]{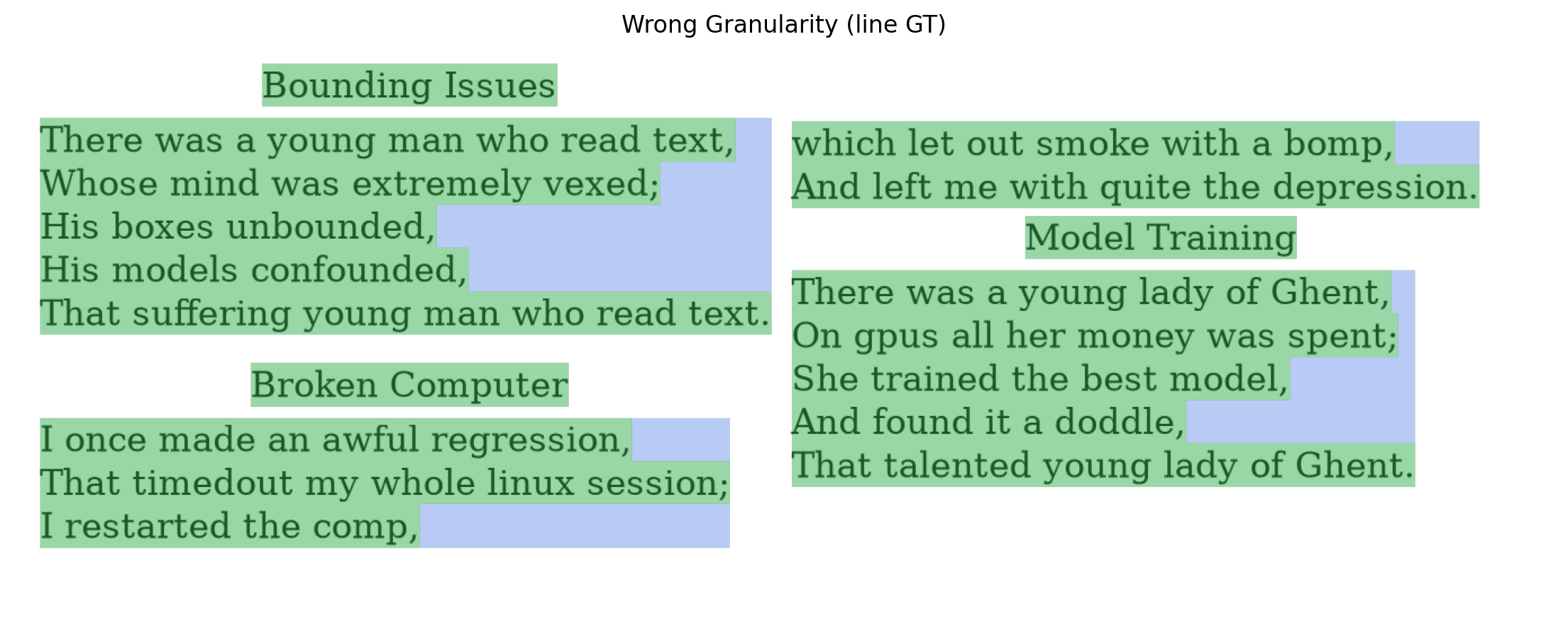}
    \caption{Granularity mismatch: GT and line level predictions at paragraph level}
  \end{subfigure}
  \hfill
  \begin{subfigure}[b]{0.49\textwidth}
    \includegraphics[width=\linewidth]{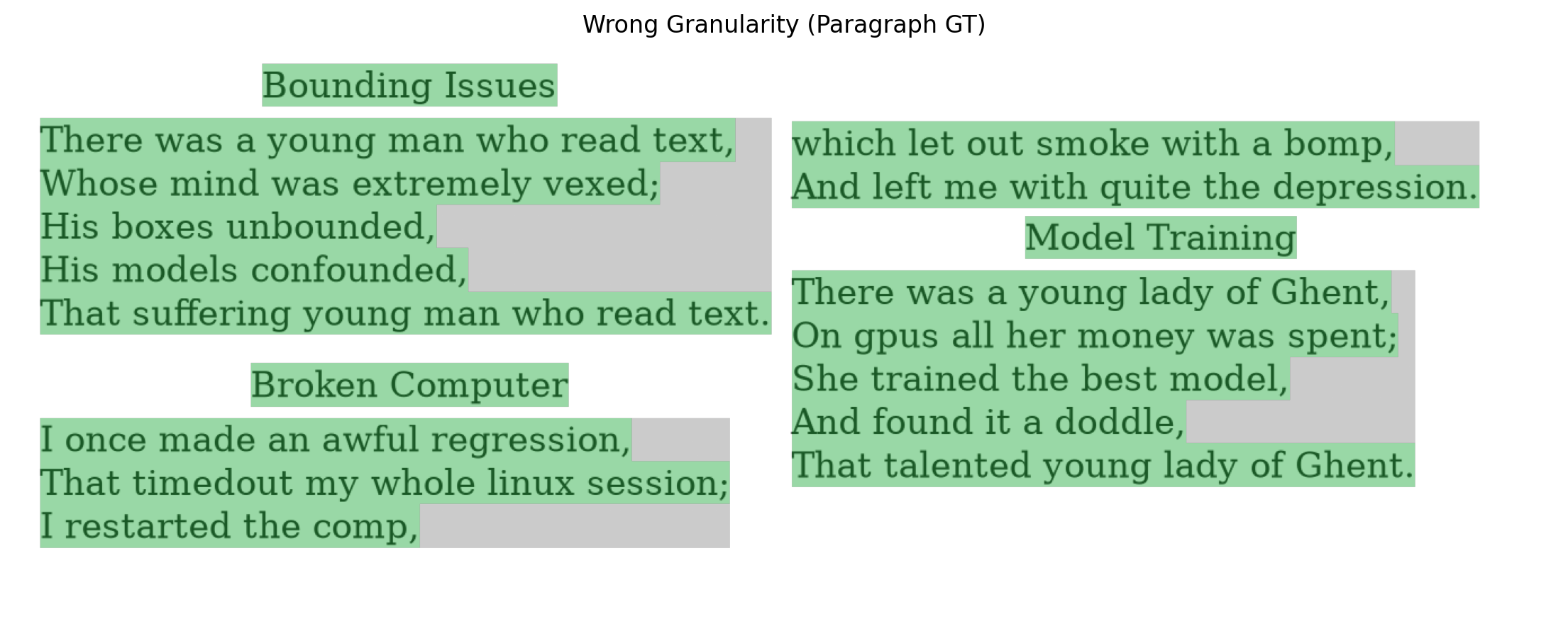}
    \caption{Granularity mismatch: GT and paragraph level predictions at line level}
  \end{subfigure}
  \caption{Different parsing issues and errors}
  \label{fig:additional_parsing_issues}
\end{figure}

Figure \ref{fig:additional_parsing_issues} shows additional parsing issues using the example limerick from the paper. The four sub-images show, missing predictions (grey), Overlapping predictions (yellow), and two types of granularity mismatch. The full MOTe error, F1 at IoU$\geq0.5$, mean IoU, for the four examples and the complex errors used in the main paper are shown in Table \ref{tab:cote_cer_table}. In addition to measured parsing errors, Table \ref{tab:cote_cer_table} also shows the observed Character Error Rate (CER) of each case under perfect text extraction. We show the CER as it demonstrates that poor parsing can produce apparent errors in OCR even when the OCR engine has faithfully transcribed the image it has been presented with. These errors appear because; in the case of complex errors example, the prediction region contains characters out of order within the ground truth box; and in the case of overlap additional characters are inserted into the final text. 

\begin{table}[h]
\centering
\begin{tabular}{lccccccc c}
\toprule
Prediction set & MOTe & $\mathcal{M}$ & $\mathcal{O}$ & $\mathcal{T}$ & e & F1@0.5 & mIoU & CER \\
\midrule
Missing regions        & 0.15 & 0.15 & 0.00 & 0.00 & 0.00 & 0.83 & 0.71 & 0.17 \\
Duplicate regions      & 0.61 & 0.00 & 0.61 & 0.00 & 0.00 & 0.87 & 1.00 & 0.59 \\
Granularity, line GT   & 0.00 & 0.00 & 0.00 & 0.00 & 0.15 & 0.32 & 0.36 & 0.00 \\
Granularity, SSU GT    & 0.16 & 0.16 & 0.00 & 0.00 & 0.00 & 0.32 & 0.61 & 0.00 \\
Complex errors       & 0.33 & 0.09 & 0.12 & 0.12 & 0.07 & 0.76 & 0.64 & 0.36 \\
\bottomrule
\end{tabular}
\caption{There is a substantial range of outcomes across the case studies and metrics demonstrating the importance of decomposable errors. We also see that parsing errors induce apparent OCR errors even when there are none.}
\label{tab:cote_cer_table}
\end{table}

\subsubsection{Class based analysis}

The example has only two classes within it `title' and `paragraph'. The MOTe F1 of the parsing is 0.91, the relative Coverage, Overlap and Trespass of the two classes are shown in Table \ref{tab:limerick_class_shares}, and are obviously dominated by the text class, which makes up the majority of the image. 

\begin{table}[htbp]
\centering
\caption{Share of total Coverage, Overlap and Trespass attributed to each predicted class.}
\label{tab:limerick_class_shares}
\begin{tabular}{lccc}
\toprule
\textbf{Class} & \textbf{$\mathcal{C}^{\textrm{share}}_k$} & \textbf{$\mathcal{O}^{\textrm{share}}_k$} & \textbf{$\mathcal{T}^{\textrm{share}}_k$} \\
\midrule
headline & 0.05 & 0.04 & 0.00 \\
text & 0.95 & 0.96 & 1.00 \\
\bottomrule
\end{tabular}
\end{table}

Tables \ref{tab:limerick_overlap_confusion} and \ref{tab:limerick_trespass_confusion} show the Overlap and Trespass confusion matrices respectively.

\begin{table}[htbp]
\centering
\caption{Overlap confusion matrix between classes. Each column is a predicted class and sums to one.}
\label{tab:limerick_overlap_confusion}
\begin{tabular}{lcc}
\toprule
\textbf{GT $\downarrow$ / Pred $\rightarrow$} & \textbf{headline} & \textbf{text} \\
\midrule
headline & 1.00 & 0.05 \\
text & 0.00 & 0.95 \\
\bottomrule
\end{tabular}
\end{table}

\begin{table}[htbp]
\centering
\caption{Trespass confusion matrix between classes. Each column is a predicted class and sums to one.}
\label{tab:limerick_trespass_confusion}
\begin{tabular}{lcc}
\toprule
\textbf{GT $\downarrow$ / Pred $\rightarrow$} & \textbf{headline} & \textbf{text} \\
\midrule
headline & 0.00 & 0.09 \\
text & 0.00 & 0.91 \\
\bottomrule
\end{tabular}
\end{table}

\section{Correlation with CER}

Table \ref{tab:grid_corr} shows the results using the controlled dataset. The table shows that MOTe has low absolute correlation on the no-noise sample, as this is dominated by vertical Trespass, which does not induce character error; with noise, MOTe becomes strongly correlated with CER. F1 is poorly correlated across all types because it struggles with granularity differences and cannot properly express Overlap and trespass, producing many false positives and failing to penalise or reward partial coverage. Dice has moderate to strong correlation across all combinations but the correlation switches sign depending if the prediction and ground truth columns match or not, making it quite unreliable. The lower bound $\max(M, O)$ has very high absolute correlation when the prediction and ground truth columns match, but it collapses to near 0 when they don't, due to Trespass causing parsing-induced character errors. 

There is clearly a large difference between the mean and pooled values for each group. This is because structural differences between the groups cause the pooled mean to track the group-mean correlation, making the mean correlation a better measure.

\begin{table}
\centering
\caption{Spearman correlation of CER with each score per GT $\times$ pred column cell and noise condition, the unweighted mean over the nine cells, and pooled over all cells of a condition. MOTe, $\max(M, O)$ and the components M, O and T are errors, so positive values indicate agreement with CER; F1@0.5 and Dice are scores, so negative values indicate agreement with CER.}
\label{tab:grid_corr}
\begin{tabular}{llrrrrrrr}
\toprule
 &  & MOTe & $\max(\mathcal{M},\mathcal{O})$ & F1@0.5 & Dice & $\mathcal{M}$ & $\mathcal{O}$ & $\mathcal{T}$ \\
\midrule
\multirow[c]{11}{*}{none} & GT 1 $\to$ pred 1 & -0.12 & 0.86 & -0.16 & -0.61 & 0.86 & -- & -0.30 \\
 & GT 1 $\to$ pred 2 & 0.36 & -0.56 & -0.22 & 0.34 & -0.56 & -- & 0.41 \\
 & GT 1 $\to$ pred 3 & 0.40 & -0.43 & -0.06 & 0.25 & -0.43 & -- & 0.42 \\
 & GT 2 $\to$ pred 1 & 0.01 & -0.51 & -0.14 & 0.53 & -0.51 & -- & 0.36 \\
 & GT 2 $\to$ pred 2 & -0.13 & 0.75 & -0.16 & -0.57 & 0.75 & -- & -0.20 \\
 & GT 2 $\to$ pred 3 & 0.06 & -0.14 & -0.01 & 0.24 & -0.14 & -- & 0.07 \\
 & GT 3 $\to$ pred 1 & -0.22 & -0.81 & -0.10 & 0.77 & -0.81 & -- & 0.74 \\
 & GT 3 $\to$ pred 2 & 0.30 & -0.16 & 0.06 & 0.09 & -0.16 & -- & 0.32 \\
 & GT 3 $\to$ pred 3 & -0.05 & 0.69 & -0.13 & -0.46 & 0.69 & -- & -0.10 \\
 & Mean of cells & 0.07 & -0.03 & -0.10 & 0.06 & -0.03 & -- & 0.19 \\
 & Pooled & 0.65 & 0.48 & -0.77 & -0.54 & 0.48 & -- & 0.61 \\
\midrule
\multirow[c]{11}{*}{within} & GT 1 $\to$ pred 1 & 0.70 & 0.95 & -0.10 & -0.47 & 0.45 & 0.31 & -0.17 \\
 & GT 1 $\to$ pred 2 & 0.50 & 0.06 & -0.02 & 0.68 & -0.70 & 0.76 & 0.56 \\
 & GT 1 $\to$ pred 3 & 0.51 & -0.03 & 0.02 & 0.73 & -0.75 & 0.77 & 0.64 \\
 & GT 2 $\to$ pred 1 & 0.62 & 0.25 & -0.02 & 0.48 & -0.48 & 0.60 & 0.61 \\
 & GT 2 $\to$ pred 2 & 0.71 & 0.96 & -0.09 & -0.45 & 0.44 & 0.38 & -0.12 \\
 & GT 2 $\to$ pred 3 & 0.62 & 0.34 & 0.04 & 0.43 & -0.44 & 0.76 & 0.42 \\
 & GT 3 $\to$ pred 1 & 0.66 & 0.08 & -0.00 & 0.64 & -0.64 & 0.69 & 0.79 \\
 & GT 3 $\to$ pred 2 & 0.59 & 0.82 & -0.06 & -0.32 & 0.31 & 0.29 & -0.06 \\
 & GT 3 $\to$ pred 3 & 0.72 & 0.95 & -0.14 & -0.44 & 0.43 & 0.42 & -0.09 \\
 & Mean of cells & 0.63 & 0.49 & -0.04 & 0.14 & -0.15 & 0.55 & 0.29 \\
 & Pooled & 0.63 & 0.27 & -0.64 & 0.01 & -0.06 & 0.29 & 0.53 \\
\midrule
\multirow[c]{11}{*}{random} & GT 1 $\to$ pred 1 & 0.70 & 0.90 & -0.19 & -0.46 & 0.44 & 0.33 & -0.07 \\
 & GT 1 $\to$ pred 2 & 0.38 & -0.03 & -0.14 & 0.61 & -0.63 & 0.54 & 0.49 \\
 & GT 1 $\to$ pred 3 & 0.30 & -0.10 & -0.22 & 0.61 & -0.62 & 0.44 & 0.49 \\
 & GT 2 $\to$ pred 1 & 0.30 & 0.13 & -0.19 & 0.34 & -0.34 & 0.19 & 0.38 \\
 & GT 2 $\to$ pred 2 & 0.72 & 0.91 & -0.18 & -0.46 & 0.43 & 0.39 & -0.01 \\
 & GT 2 $\to$ pred 3 & 0.56 & 0.33 & -0.04 & 0.36 & -0.37 & 0.59 & 0.37 \\
 & GT 3 $\to$ pred 1 & 0.36 & 0.00 & -0.23 & 0.49 & -0.49 & 0.28 & 0.58 \\
 & GT 3 $\to$ pred 2 & 0.46 & 0.67 & -0.20 & -0.32 & 0.30 & 0.09 & -0.04 \\
 & GT 3 $\to$ pred 3 & 0.75 & 0.90 & -0.22 & -0.47 & 0.42 & 0.43 & 0.09 \\
 & Mean of cells & 0.50 & 0.41 & -0.18 & 0.08 & -0.10 & 0.36 & 0.25 \\
 & Pooled & 0.53 & 0.26 & -0.57 & -0.00 & -0.04 & 0.21 & 0.43 \\
\bottomrule
\end{tabular}
\end{table}

We see in Table \ref{tab:grid_weighting_diagonal} that $\max(M, O)$ works well on the diagonal whilst $M + O + 2T$ works reasonably well in both scenarios when there is noise; the basic MOTe works well on and off the diagonal when there is noise. This appears related to the difference between horizontal and vertical Trespass, which is supported by the findings of Table \ref{tab:real_weighting_columns}, which show that $\max(M, O)$ has the highest absolute correlation, as the models generally made very little horizontal Trespass relative to the vertical Trespass. The tension between these results suggests that while $\max(M, O)$ performs better on real data, it has a major weakness when horizontal Trespass becomes more prevalent. The choice of metric is therefore a trade-off, with MOTe as the best all-rounder.

\begin{table}
\centering
\caption{Mean Spearman correlation of CER with MOTe and its re-weightings over the diagonal (GT and pred column counts match) and off-diagonal cells, per noise condition. All three are errors, so positive values indicate agreement with CER.}
\label{tab:grid_weighting_diagonal}
\begin{tabular}{llrrr}
\toprule
 &  & MOTe & $\max(M, O)$ & $M + O + 2T$ \\
Noise & Diagonal &  &  &  \\
\midrule
\multirow[c]{2}{*}{none} & on & -0.10 & 0.77 & -0.15 \\
 & off & 0.15 & -0.43 & 0.34 \\
\midrule
\multirow[c]{2}{*}{within} & on & 0.71 & 0.95 & 0.41 \\
 & off & 0.58 & 0.25 & 0.56 \\
\midrule
\multirow[c]{2}{*}{random} & on & 0.72 & 0.90 & 0.46 \\
 & off & 0.39 & 0.17 & 0.40 \\
\bottomrule
\end{tabular}
\end{table}

\begin{table}
\centering
\caption{Spearman correlation of CER with MOTe and its trespass re-weightings for real layout models, within pages of each GT column count and the unweighted mean over the three. Pooled stacks all models. All three are errors, so positive values indicate agreement with CER.}
\label{tab:real_weighting_columns}
\begin{tabular}{llrrr}
\toprule
 &  & MOTe & $\max(\mathcal{M},\mathcal{O})$ & $\mathcal{M} + \mathcal{O} + 2\mathcal{T}$ \\
Model & GT columns &  &  &  \\
\midrule
\multirow[c]{4}{*}{Pooled} & 1 & 0.64 & 0.94 & 0.54 \\
 & 2 & 0.87 & 0.92 & 0.77 \\
 & 3 & 0.87 & 0.82 & 0.72 \\
 & Mean & 0.79 & 0.89 & 0.67 \\
\bottomrule
\end{tabular}
\end{table}

\iffalse
Table \ref{tab:real_cote_cer_correlation}, shows the break down of correlation scores across the F1 the MOTe, and its decomposed elements. What is notable is that Missing changes sign between GT-prediction pairs when the share the same number of columns and when they don't. This is because coverage broadly aligns with the SSU structure when the columns are the same meaning less Missing on average reduces CER. In contrast when the number of columns are different, less Missing produces more Trespass, or breaks the assumption about reading order being internally recoverable. 

\begin{table}
\centering
\caption{Spearman correlation of CER with each score for real layout models on the synthetic pages.}
\label{tab:real_cote_cer_correlation}
\begin{tabular}{lrrrrrrrrr}
\toprule
 & n & MOTe & $\max(\mathcal{M}, \mathcal{O})$ & F1@0.5 & Dice & $\mathcal{M}$ & $\mathcal{O}$ & $\mathcal{T}$ & E \\
\midrule
Pooled & 2250 & 0.82 & 0.89 & -0.54 & -0.62 & 0.41 & 0.31 & 0.13 & -0.05 \\
DocLayout-YOLO & 450 & 0.91 & 0.98 & -0.55 & -0.48 & 0.24 & 0.95 & 0.56 & 0.28 \\
DoclingLayoutHeron & 450 & 0.81 & 0.88 & -0.46 & -0.33 & -0.04 & 0.87 & 0.59 & 0.37 \\
PPDocLayout-L & 450 & 0.82 & 0.96 & -0.49 & -0.58 & 0.42 & 0.38 & 0.10 & -0.05 \\
PPDocLayout-M & 450 & 0.81 & 0.80 & -0.59 & -0.73 & 0.67 & -0.16 & -0.18 & -0.37 \\
PPDocLayout-S & 450 & 0.75 & 0.82 & -0.43 & -0.59 & 0.44 & 0.14 & -0.08 & -0.29 \\
\bottomrule
\end{tabular}
\end{table}

\fi

\section{Case Study: MOTe and natural scenes}
Although the MOTe error was designed to analyse model parsing quality of documents, it also functions as a general-purpose machine vision evaluator, fulfilling a similar but different role to Panoptic Quality (PQ) \cite{kirillov_panoptic_2019}. PQ is an IoU based metric designed for panoptic segmentation. Panoptic Segmentation is a machine vision approach similar to semantic segmentation in that each pixel of the image is labelled as a single class. However, unlike basic segmentation, panoptic segmentation distinguishes between objects of the same class; as such, multiple pixels may be tagged as ``car" but are considered distinct cars. Like the MOTe error, Panoptic Quality is decomposable and is calculated as

\begin{equation}
\mathrm{PQ} = \frac{\sum_{(p,g) \in \mathit{TP}} \mathrm{IoU}(p,g)}
{|\mathit{TP}| + \frac{1}{2}|\mathit{FP}| + \frac{1}{2}|\mathit{FN}|}
\label{eq:pq}
\end{equation}

\begin{equation}
\mathrm{PQ} =
\underbrace{\frac{\sum_{(p,g) \in \mathit{TP}} \mathrm{IoU}(p,g)}{|\mathit{TP}|}}_{\text{segmentation quality (SQ)}}
\times
\underbrace{\frac{|\mathit{TP}|}{|\mathit{TP}| + \frac{1}{2}|\mathit{FP}| + \frac{1}{2}|\mathit{FN}|}}_{\text{recognition quality (RQ)}}
\label{eq:pq_decomposed}
\end{equation}

\begin{equation}
\mathrm{IoU}(p,g) = \frac{|p \cap g|}{|p \cup g|}
\label{eq:iou}
\end{equation}

Using the same notation as the F1 from the main manuscript. The decomposition is the Recognition Quality and the Segmentation Quality. Recognition Quality is the F1, whilst Segmentation Quality provides a mechanism for understanding the parse quality in terms of the IoU.

This section provides a simple demonstration example of the MOTe error applied to natural scenes and contrasts it with the PQ. We use two images from the COCO dataset \cite{lin_microsoft_2015}, and two off-the-shelf models, Yolo26-instance  \cite{jocher_ultralytics_2026}, and Mask2Former \cite{cheng_masked-attention_2022}. Segmentation tasks label all pixels and divide the image into roughly two categories: Stuff (background objects, like, grass, sky, stones), and things (people, dogs, bicycles). Mask2Former, as a segmentation model, handles this naturally; however, Yolo26 is an instance model and handles only Things. For fairness, we compare the models on only the Things in the image.

\begin{figure}[htbp]
  \centering
  \begin{subfigure}[b]{0.49\textwidth}
    \includegraphics[width=\linewidth]{supplementary/coco_elephant_325871.jpg}
    \caption{Raw image}
    \label{fig:a_ele}
  \end{subfigure}
  \hfill
  \begin{subfigure}[b]{0.49\textwidth}
    \includegraphics[width=\linewidth]{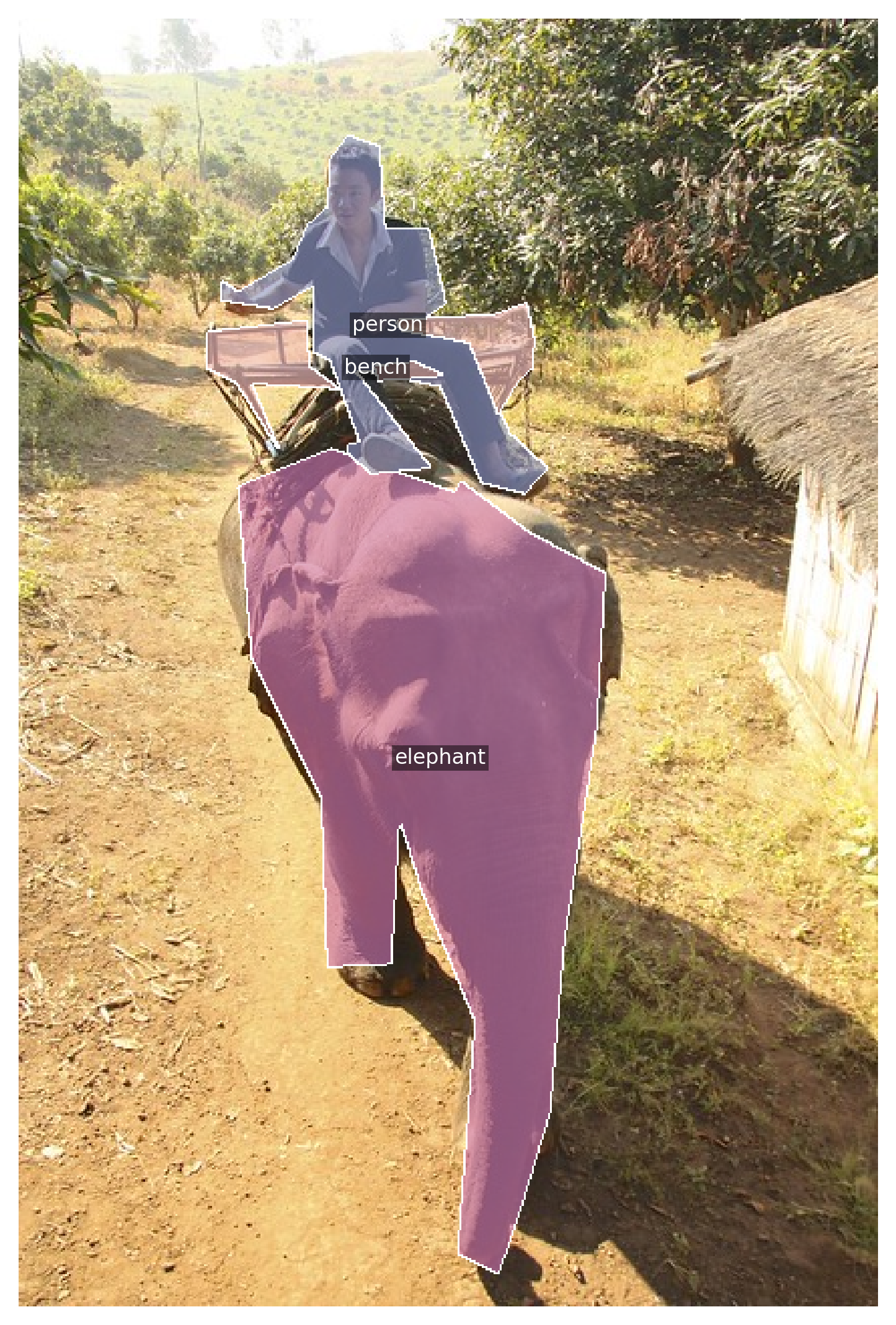}
    \caption{Ground Truth masks}
    \label{fig:b_ele}
  \end{subfigure}

  \vspace{0.5em}

  \begin{subfigure}[b]{0.49\textwidth}
    \includegraphics[width=\linewidth]{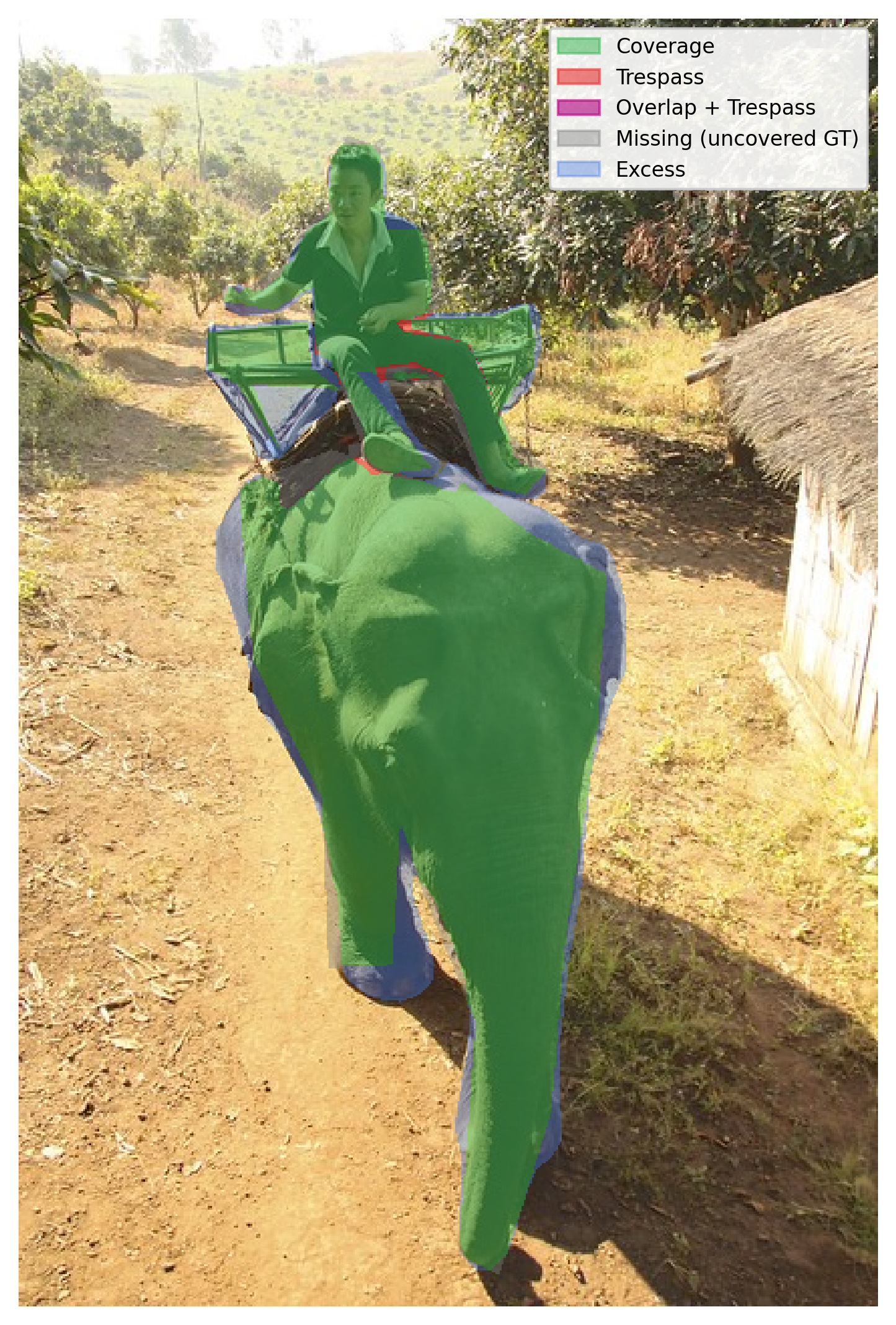}
    \caption{Yolo predictions}
  \end{subfigure}
  \hfill
  \begin{subfigure}[b]{0.49\textwidth}
    \includegraphics[width=\linewidth]{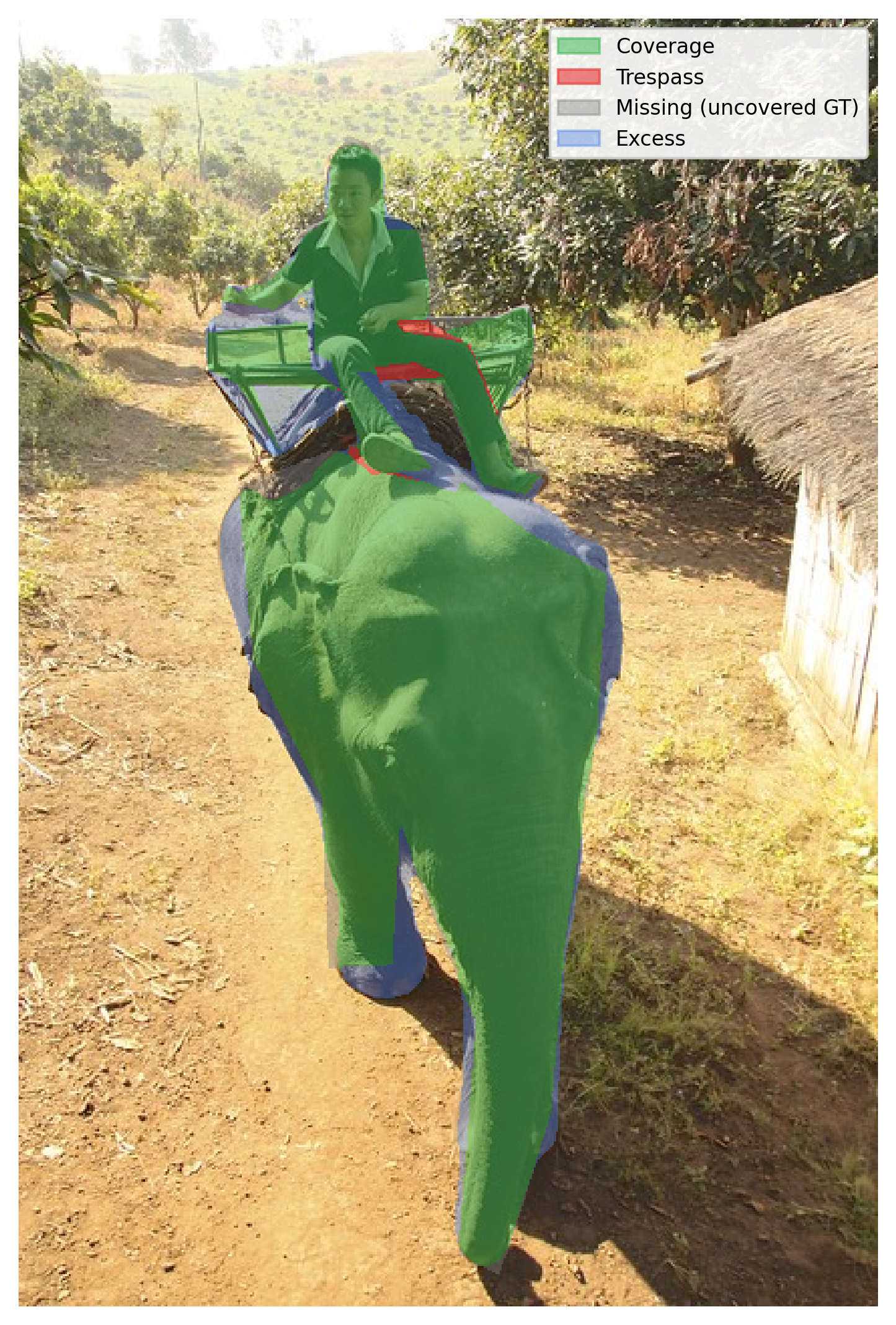}
    \caption{Mask2Former predictions}
  \end{subfigure}
  \caption{Man sitting on howdah and riding an elephant.}
  \label{fig:coco_ele}
\end{figure}

\begin{figure}[htbp]
  \centering
  \begin{subfigure}[b]{0.49\textwidth}
    \includegraphics[width=\linewidth]{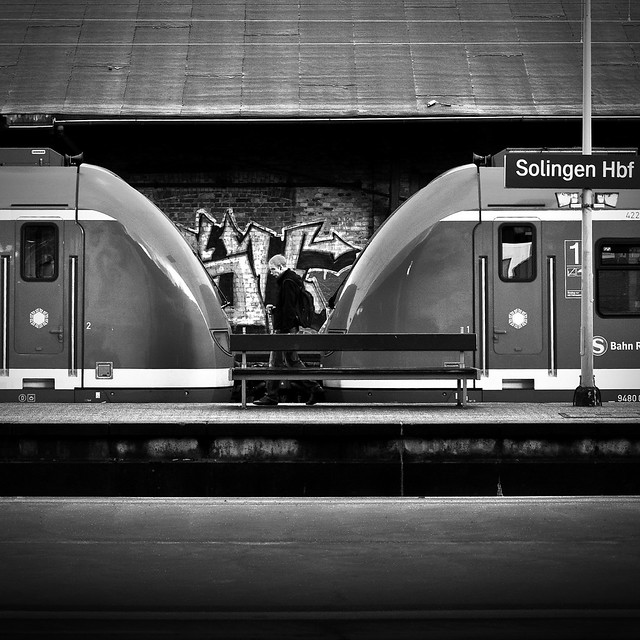}
    \caption{Raw image}
    \label{fig:a_train}
  \end{subfigure}
  \hfill
  \begin{subfigure}[b]{0.49\textwidth}
    \includegraphics[width=\linewidth]{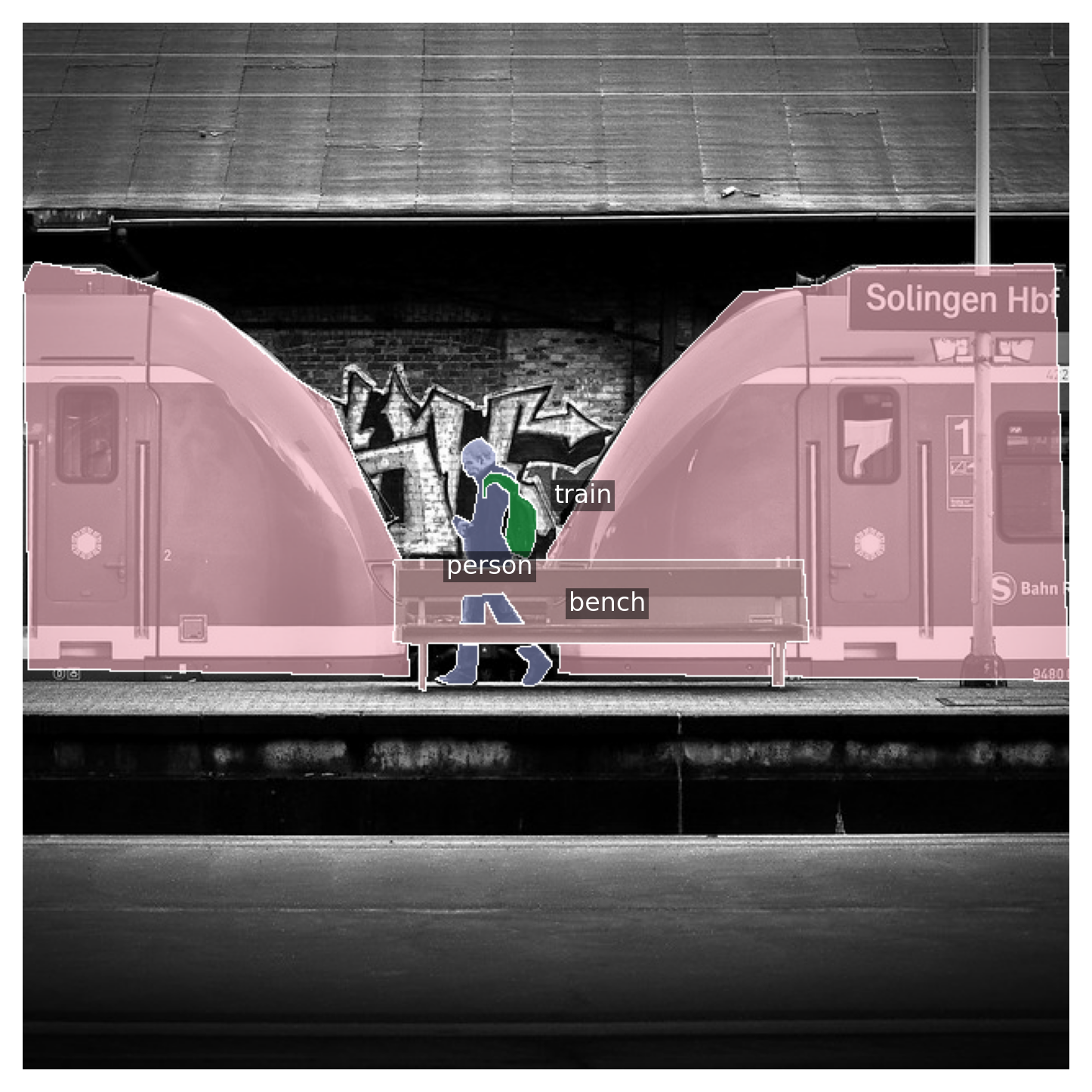}
    \caption{Ground Truth masks}
    \label{fig:b_train}
  \end{subfigure}

  \vspace{0.5em}

  \begin{subfigure}[b]{0.49\textwidth}
    \includegraphics[width=\linewidth]{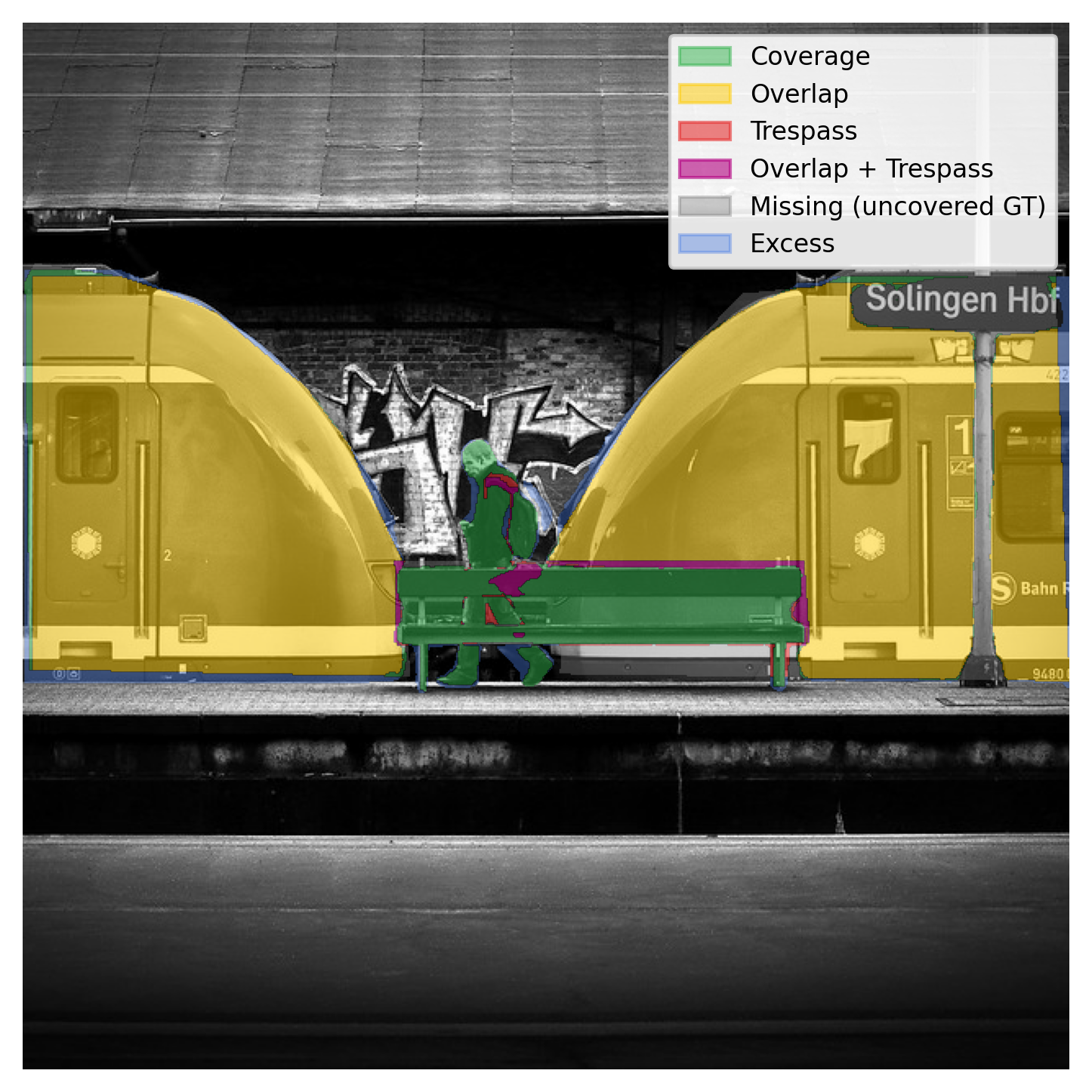}
    \caption{Yolo predictions}
  \end{subfigure}
  \hfill
  \begin{subfigure}[b]{0.49\textwidth}
    \includegraphics[width=\linewidth]{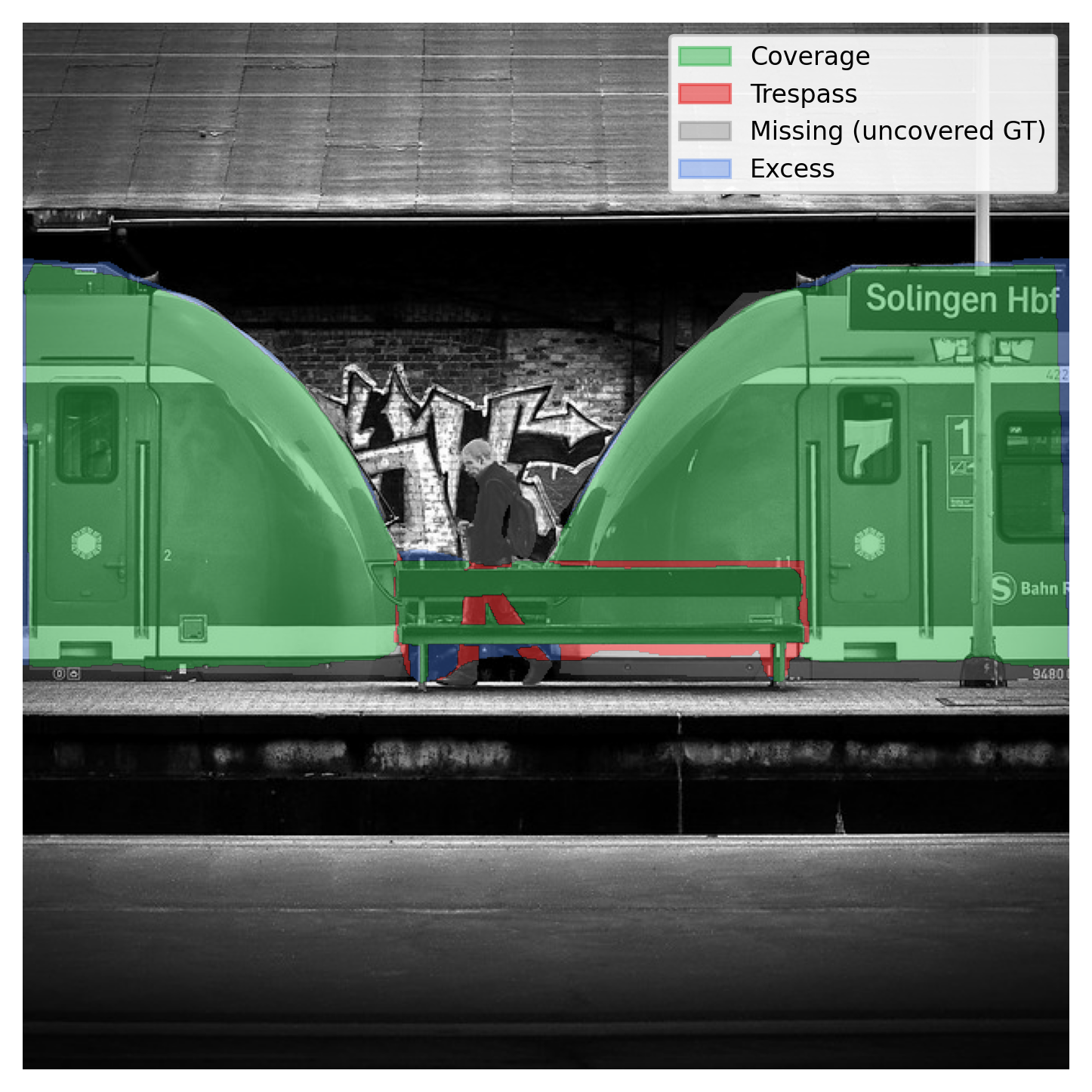}
    \caption{Mask2Former predictions}
  \end{subfigure}
  \caption{Image of a train station}
  \label{fig:coco_train}
\end{figure}

Figures \ref{fig:coco_ele} and \ref{fig:coco_train} each show, four sub-images, the raw image, the GT, prediction using Yolo26 \cite{jocher_ultralytics_2026}, and prediction using Mask2Former \cite{cheng_masked-attention_2022}.

We can see from Figure \ref{fig:coco_ele} that both models perform well. Indeed the excess they create is not due to poor mask creation but to the poor quality of the ground truth that produces a jagged outline on both the Elephant and the rider. Table \ref{tab:coco_elephant} quantifies performance and shows strong performance across both MOTe and PQ, where PQ is the macro-average across classes. PQ is reduced by the IoU in the SQ component which is partly due to poor GT quality, and this is also reflected in MOTe's Excess score.

In the train image shown in Figure \ref{fig:coco_train}, MOTe and PQ diverge more substantially. Mask2Former has a much lower MOTe error than Yolo26; because Yolo double-labels the two train components: the trains are labelled as separate instances and then again as a single train instance. This causes very high overlap, bringing the overall error to 0.89 compared to Mask2Former’s 0.09. On the other hand, PQ penalises the models for different reasons. Mask2Former misses the person and backpack entirely; because PQ averages over classes, these two small objects (3\% of the thing pixels) account for half the score, while MOTe's Missing rises only to 0.07. Yolo26 finds every object, but none of its train masks reaches the IoU $>0.5$ match threshold against the single ground-truth train, so the class scores zero, whereas Mask2Former's right-hand train clears it at 0.51, partly because, like the ground truth, it includes the station sign, which Yolo26 correctly does not include as part of the train. It is worth noting that the ground truth marks both the right and left train as a single object; however, there is no visual evidence in the image to suggest this is true as opposed to two trains nose-to-nose in the station. These differences in the way the metrics evaluate the output of the models, and the way the ground truth is labelled, can all be valid, but depend on the goals of the modeller and the downstream requirements.

What is interesting is that the MOTe F1 is high across both images for both models, even for Yolo26 on the train. This is because the MOTe F1 measures whether a pixel was assigned the correct class independent of instance and ignoring overlap. It is functionally looking at the segmentation of the image. 

From this simple example we can see that there are clear design differences between the MOTe and PQ, this combined with the issues noted on the ground truth,  brings us back to the meaning-as-use discussion from the main paper, raising the same questions of a one-size-fits-all approach to machine learning.

An interesting question for further work is to properly analyse the MOTe applied to natural scenes, the value of the decomposition, and the limitations of not distinguishing between instances within the image compared with the PQ. Furthermore, it is an interesting question to ask what the impact of replacing the Segmentation Quality (SQ) with $1 - \mathrm{MOTe}$ would be in the ability of the PQ to provide insight into model performance in natural scenes. If such a swap were made, the overall score would still be limited by the Recognition Quality/F1 term, but segmentation would penalise Overlap and Trespass.

\begin{table}[htbp]
\centering
\caption{MOTe and panoptic quality on the COCO elephant image (things scope). MOTe, Missing, Overlap, Trespass and Excess are lower-is-better; MOTe F1, PQ, SQ and RQ are higher-is-better.}
\label{tab:coco_elephant}
\begin{tabular}{lccccccccc}
\toprule
\textbf{Model} & \textbf{MOTe} & \textbf{Missing} & \textbf{Overlap} & \textbf{Trespass} & \textbf{Excess} & \textbf{MOTe F1} & \textbf{PQ} & \textbf{SQ} & \textbf{RQ} \\
\midrule
Mask2Former & \textbf{0.04} & \textbf{0.03} & \textbf{0.00} & 0.01 & \textbf{0.03} & \textbf{0.91} & 0.76 & 0.76 & \textbf{1.00} \\
Yolo26 & 0.04 & 0.04 & 0.00 & \textbf{0.00} & 0.04 & 0.90 & \textbf{0.78} & \textbf{0.78} & \textbf{1.00} \\
\bottomrule
\end{tabular}
\end{table}

\begin{table}[htbp]
\centering
\caption{MOTe and panoptic quality on the COCO train image (things scope). MOTe, Missing, Overlap, Trespass and Excess are lower-is-better; MOTe F1, PQ, SQ and RQ are higher-is-better.}
\label{tab:coco_train}
\begin{tabular}{lccccccccc}
\toprule
\textbf{Model} & \textbf{MOTe} & \textbf{Missing} & \textbf{Overlap} & \textbf{Trespass} & \textbf{Excess} & \textbf{MOTe F1} & \textbf{PQ} & \textbf{SQ} & \textbf{RQ} \\
\midrule
Mask2Former & \textbf{0.09} & \textbf{0.07} & \textbf{0.00} & 0.02 & \textbf{0.01} & \textbf{0.93} & 0.27 & 0.32 & 0.42 \\
Yolo26 & 0.89 & 0.08 & 0.78 & \textbf{0.02} & 0.02 & 0.92 & \textbf{0.58} & \textbf{0.58} & \textbf{0.75} \\
\bottomrule
\end{tabular}
\end{table}

\end{document}